\documentclass[11pt]{article}
\usepackage{url}       
\usepackage{nicefrac}  
\usepackage{booktabs} 
\usepackage{amsmath}
\usepackage{amssymb}
\usepackage{amsthm}
\usepackage{amscd}
\usepackage{amsfonts}
\usepackage{dsfont}
\usepackage{graphicx}%
\usepackage{subcaption}
\usepackage{fancyhdr}
\usepackage{mathtools}
\usepackage{empheq}
\usepackage{bibunits}

\usepackage{cite}
\usepackage{xcolor}
\usepackage{algorithm}
\usepackage[algo2e,ruled]{algorithm2e}
\usepackage{ifthen}
\usepackage{algorithmic}
\usepackage{textcomp}

\newtheorem{theorem}{Theorem}
\newtheorem{lemma}{Lemma}
\newtheorem{remark}{Remark}

\newtheorem{corollary}{Corollary}

\DeclareMathOperator*{\argmin}{arg\,min}
\DeclareMathOperator*{\argmax}{arg\,max}

\SetCommentSty{mycommfont}
\def\scr#1{{\cal #1}} 
\def\eq#1{\begin{equation}#1\end{equation}}
\newcommand{\R}{{\rm I\!R}}
\newcommand{\bbb}{\mathbb}

\def\rep#1{(\ref{#1})}

\usepackage{hyperref}
\newcommand{\1}{\mathbf{1}}

\def\qed{ \rule{.08in}{.08in}}
\usepackage{parskip}
\begin{document}

\title{Decentralized Upper Confidence Bound Algorithms for Homogeneous Multi-Agent Multi-Armed Bandits}


\author{Jingxuan Zhu, 
Ethan Mulle, Christopher S. Smith, Alec Koppel, Ji~Liu\thanks{J.~Zhu is currently with E-Surfing Digital Life Technology Co., Ltd. and was previously affiliated with the Department of Applied Mathematics and Statistics at Stony Brook University. 
The majority of the work was completed while J. Zhu was at Stony Brook University.
E.~Mulle is with the Department of Applied Mathematics at University of California Santa Cruz (\texttt{emulle@ucsc.edu}).
C.S.~Smith is with the Department of Computer Science at Stony Brook University (\texttt{christopher.s.smith@stonybrook.edu}).
A.~Koppel is with J.P. Morgan AI Research (\texttt{alec.koppel@jpmchase.com}).
J.~Liu is with the Department of Electrical and Computer Engineering at Stony Brook University
(\texttt{ji.liu@stonybrook.edu}).}
}
\maketitle


\vspace{-.1in}

\begin{abstract}
This paper studies a decentralized homogeneous multi-armed bandit problem in a multi-agent network. The problem is simultaneously solved by $N$ agents assuming they face a common set of $M$ arms and share the same arms' reward distributions. Each agent can receive information only from its neighbors, where the neighbor relationships among the agents are described by a fixed graph. 
Two fully decentralized upper confidence bound (UCB) algorithms are proposed  for undirected graphs, respectively based on the classic algorithm and the state-of-the-art Kullback-Leibler upper confidence bound (KL-UCB) algorithm. The proposed decentralized UCB1 and KL-UCB algorithms permit each agent in the network to achieve a better logarithmic asymptotic regret than their single-agent counterparts, provided that the agent has at least one neighbor, and the more neighbors an agent has, the better regret it will have, meaning that the sum is more than its component parts. The same algorithm design framework is also extended to directed graphs through the design of a variant of the decentralized UCB1 algorithm, which outperforms the single-agent UCB1 algorithm.
\end{abstract}




\section{Introduction}\label{sec:intro}

Multi-armed bandit (MAB) is a fundamental framework for reasoning about incentives which are sequentially revealed \cite{lattimore2020bandit},
which has a wide range of applications in natural and engineered systems including clinical trials \cite{mason2022experimental}, adaptive routing \cite{talebi2017stochastic}, cognitive radio networks \cite{avner2014concurrent}, and online recommendation systems \cite{mary2015bandits}.  In the standard setup, a single agent makes a sequential decision to select one arm at each discrete time from a given finite set of arms and then receives a reward informing the merit of a chosen arm, generated according to a random variable with an unknown distribution. The target of the decision maker is to minimize its cumulative expected regret, i.e., the difference between the decision maker's accumulated (expected) reward and that associated with the best decision in hindsight which has access to all information in advance. Both lower and upper bounds on the asymptotic regret have been derived in the seminal work \cite{LR85}, classic UCB algorithms were proposed in \cite{auer2002finite} which achieve an $O(\log T)$ regret, and the state-of-the-art algorithm, KL-UCB, 
was crafted in \cite{KLUCB}. For an introductory tutorial, see a recent monograph \cite{survey}.

Decision-making in networked settings has gained increasing salience due to the growth in scale of social, communication, data, and transportation networks. Key axes of comparison are whether agents are cooperative or competitive \cite{zhang2021multi}, and how information exchange is executed \cite{krishnamurthy2009partially}. Our focus is on cooperative settings, due to its ability encapsulate decentralized and parallel computing architectures, sensor networks, and other applications involving interconnected devices \cite{xia2020multi} such as Internet of Things and cyber-physical systems.
While information propagation across a network can be achieved by imposing constraints that can be solved with a variety of optimization techniques \cite{yang2019survey}, the simplest possible methodology is based off agents evaluating a weighted average of their local parameters with their neighbors, i.e., the consensus protocol \cite{flocking,MurraySurvey}. The gold standard of cooperative multi-agent learning is to design a scheme such that the team is \emph{more} than the sum of its component parts. Cooperative multi-agent bandits was first proposed for the case that all agents share the same reward distribution of each arm \cite{Landgrenecc}, which has attracted numerous follow-on works  \cite{Landgrenecc,leonard,princetonauto,martinez2019decentralized,jingxuan,sigmetrics,chawla}. 

In the case that agents' rewards are drawn from homogeneous distributions, each agent in a network can independently learn an optimal arm using any conventional single-agent UCB algorithm, ignoring any information received from other agents. Notwithstanding, 
all existing algorithms, except for that in \cite{jingxuan2021}, for the decentralized multi-armed bandit problem with homogeneous reward distributions 
require that each agent be aware of certain network-wide global information, such as spectral properties of the underlying graph, the total number of agents in the network, or the global ordering of agents (i.e., each agent has a unique identification number). 
This gap emerges due to a prior lack of explicit connection between algorithm design and quantitative relations among information fusion weights, variance estimates, and local upper confidence bound functions. Such a connection is key to existing analyses of information propagation methodologies in the supervised learning setting -- see \cite{nedic2009distributed,duchi2011dual,shi2015extra}, for instance.
That these algorithms require statistics that hinge upon global information imposes restrictions relative to the single agent case: each agent in a multi-agent network can collect more arm-related information, but its learning is not boosted from the information of others, and its learning is more restrictive.
Indeed, if a decentralized bandit algorithm does not outperform its single-agent counterpart, there is no motivation for agents to collaborate. 

Even though \cite{jingxuan2021} shows collaborating with neighbors can improve regret bounds compared with the classic single-agent UCB1, it does not exhibit any specific dependence on the network or neighborhood, meaning the incentive for collaboration is not reflected in the learning rates. That is, a neighborhood-independent improvement fails to capture the gain an agent should experience if it has more neighbors with which to communicate. Intuitively, a lower regret should be obtainable when more neighbors are available. Aside from these theoretical issues, on the algorithmic side, due to the requirement of global knowledge of network-dependent parameters in prior works \cite{Landgrenecc,leonard,princetonauto,martinez2019decentralized,jingxuan,sigmetrics,chawla}, to our knowledge there is no (fully) decentralized version for state-of-the-art bandit algorithms. 
With these in mind, we pose the following question:

{\em Can we design fully decentralized bandit algorithms, both classic and state-of-the-art ones, for the homogeneous setting without using any network-wide information, which outperform their single-agent counterparts with improved regret bounds reflecting local network effects?}

This paper provides a positive answer to this question by establishing a fully decentralized design framework for UCB algorithms, which leads to two decentralized bandit algorithms, based on the classic UCB1 and the state-of-the-art KL-UCB, respectively. Both guarantee that each agent in a network strictly outperforms their single-agent counterparts, as long as the agent has at least one neighbor. We also show that the more neighbors an agent has, the better regret bound it will have. 
We further extend the results of the decentralized UCB1 algorithm to encompass more general, directed, strongly connected graphs.





\textbf{Technical Challenges:}
We aim to propose a fully decentralized algorithm design for the decentralized multi-agent MAB problem that is compatible with different UCB algorithms. More specifically, we seek to design a common algorithm structure and updating rule for different decentralized UCB algorithms and follow a similar approach to design the algorithm-specific parameter required (usually known as the upper confidence bound).

It is well-known that one of the bottlenecks of bandit problems is an agent's insufficient exploration on some arm, which consequently may cause it to keep pulling a sub-optimal arm, leading to a linear regret. Single-agent UCB algorithms have therefore been proposed to solve this issue. Various UCB algorithms have been proposed in the literature~\cite{auer2002finite,AUDIBERT20091876,JMLR:v11:audibert10a,KLUCB}, all of which use different decision-making protocols. Based on the single-agent algorithms, we manage to construct a common structure for a decentralized version of both KL-UCB and UCB1.

The decision-making step for decentralized multi-agent setting is much more complicated than the single-agent case. This is because in the decentralized multi-agent setting, agents constantly exchange information in their neighborhood. If one agent fails to make reasonable estimation on some arm due to its relatively insufficient exploration compared with other agents, the inaccuracy it causes may be ``contagious'' to the estimation of all the agents via information fusion over the network and thus gets amplified over time. That is to say, the accuracy of estimation and its convergence rate may be dramatically affected if the exploration processes on each arm are not kept ``on the same page''. 
Another salient challenge arises due to information latency: although each agent can directly or indirectly receive processed information from all other agents in the same connected component, it takes extra time from the agents other than its neighbors. Thus, the information each agent receives does not reveal the ``current'' states of all the other agents. 
These challenges become more significant when communication between neighboring agents is uni-directional; in other words, the underlying neighbor graphs are directed.
Meanwhile, each agent may have different exploration trajectories of the arms. Information coupling and latency may further increase this exploration ``imbalance'' among the network.
To tackle this, we add a bifurcation point in the decision making step of the algorithm for each agent so as to ensure none of the agent falls too far behind in exploring each arm. More detailed clarification can be found in ``Key Design Ideas''. In addition, by saying ``fully decentralized'', agents are not aware of any network-wide information (e.g., spectral properties of the underlying graph, the network size, the unique ID number of each agent), which is required in nearly all the existing multi-agent MAB algorithms~\cite{Landgrenecc,leonard,princetonauto,martinez2019decentralized,jingxuan,contextual}, and agents are not allowed to transmit any ``raw'' information (e.g., reward samples) directly due to  immediate privacy-leaking concerns.\footnote{Agents do not want to directly share their rewards due to lower-level privacy concerns, which occur in scenarios like multi-player games and social networks. How to effectively counter a sophisticated, dedicated effort to discover agents' rewards is another matter and beyond the scope of this paper.} 
This not only restricts the type of information allowed to be transmitted, but also affects the parameter design and adaptive modification on the algorithm-wise decision making protocol of each single-agent UCB algorithm. We manage to bound the level of variation of agent's reward mean estimation, which is proved to be determining for the parameter design in the decision making protocol by combining the properties of the common updating rule and the specific techniques used for different single-agent algorithms.

\textbf{Contributions:}
The contributions of this paper are four-fold. (1) We develop novel decentralized design and analysis techniques for crafting fully decentralized UCB algorithms to solve the homogeneous multi-agent MAB problems on undirected graphs, without requiring any global information. These techniques lead to, and are demonstrated by, the following contributions. 
(2) We design the first fully decentralized KL-UCB algorithm over undirected graphs, which makes use of the state-of-the-art bandit technique and thus outperforms all existing decentralized UCB algorithms. (3) We propose a new fully decentralized UCB1 algorithm for undirected graphs, which outperforms the existing one. 
(4) 
Both our algorithms are shown to guarantee each agent to achieve a better logarithmic asymptotic regret than the single-agent counterpart, provided the agent has at least one neighbor. More importantly, we further show that the more neighbors an agent has, the better regret it will have, which well incentives collaboration among neighboring agents in a multi-agent bandit network.
(5) We extend the fully decentralized UCB1 algorithm to directed, possibly unbalanced, strongly connected graphs, which still outperforms the single-agent UCB1.

The four main contributions together provide a confirmative and comprehensive answer to the earlier posed research question.
As the key technique of UCB algorithms is to apply specific decision making protocols to the reward mean estimate, and the parameter used in the decision making protocols can be obtained by combining the properties of the common updating rule and the techniques of each single-agent UCB algorithm, we expect the decentralized algorithm design is compatible with most, if not all, UCB algorithms after adaptive modification on case (a) of the decision making step of in Section~\ref{sec:algorithm}. 

Our decentralized design and analysis techniques for UCB1 are novel and more ``advanced'' than \cite{jingxuan2021}, thus leading to a more general decentralized algorithm with superior regret compared to that in \cite{jingxuan2021}. First, the decentralized UCB1 algorithm design here can cope with directed, strongly connected graphs (cf. Theorem \ref{thm:directed}), whereas the algorithm in \cite{jingxuan2021} only works for undirected graphs. Second, the asymptotic regret bound of each agent using the algorithm in \cite{jingxuan2021} is $\frac{16}{3\Delta_k}\log T$; although better than that of the single agent UCB1, $\frac{8}{\Delta_k}\log T$, the constant coefficient improvement does not reflect any network/neighborhood effect. 
We exactly solve this issue. Our new decentralized UCB1 algorithm achieves $O(\log T/|\scr N_i|)$ asymptotic regret bound, which shows that the more neighbors an agent has, the better regret bound it will have. In addition, in the case when an agent can communicate with all other agents (e.g. a complete graph), the bound will be $O(\log T/ N)$, which is best possible regret because the lower bound of any agent's regret in a decentralized algorithm is $\Omega(\log T/N)$ (see Remark 4). To achieve such a neighborhood-dependent and best possible regret bound, more delicate algorithm design 
and complicated analysis are needed. Advantage of our algorithm over \cite{jingxuan2021} is illustrated by the simulations in ``Algorithm Comparison'' part in Section~\ref{appendix:simulation}.

\textbf{Related Work:} 
Since UCB1 and UCB2 were proposed in \cite{auer2002finite} to solve decision making bandit problems, various additional UCB algorithms have been designed, such as UCB-V~\cite{AUDIBERT20091876}, which uses an empirical Bernstein bound to obtain the retained upper confidence bound, MOSS~\cite{JMLR:v11:audibert10a}, whose optimal rate is independent of distribution, and KL-UCB~\cite{KLUCB}, which applies the Bernoulli KL-divergence to the decision-making protocol design. In recent years, as multi-agent MAB problems have been widely studied in various settings~\cite{5074370,liu2010distributed,2013,kalathil2014decentralized,ilai2018distributedbandit,sankararaman2019social,Wang2020Distributed,sandy,aaai,princetonecc,princetonletter,Landgrenecc,martinez2019decentralized}, such as ``collision''~\cite{jain,liu2010distributed,ilai2018distributedbandit,5074370} and federated settings~\cite{aaai,federated_personalized,sigmetrics, sandy,9174297,reda:hal-03825099}, consensus-based decentralized UCB algorithms have been developed in \cite{jingxuan, sigmetrics, Landgrenecc,leonard,princetonauto,martinez2019decentralized,jingxuan2021} for homogeneous reward distributions in a cooperative multi-agent setting. 
Very recently, cooperative multi-agent UCB has been extended to heterogeneous reward settings, that is, different agents may have different reward distributions and means for each arm. A heterogeneous decentralized problem is solved in \cite{sigmetrics} using the idea of gossiping to improve communication efficiency and privacy. 
All these consensus- or gossip-based algorithms, with the exception of \cite{jingxuan2021}, require an agent's awareness of certain global information, such as agent number or a certain graph parameter. Moreover, their designs are all based on single-agent UCB1 only, and are thus not general enough. 
To our knowledge, there is no existing decentralized KL-UCB algorithm. 

\section{Problem Formulation}\label{sec:problemformulate}

Consider a network consisting of $N$ agents (or players).
For presentation purposes, we label the agents from $1$ through $N$, and denote the set of agents by $[N]\triangleq\{1,2,\ldots,N\}$.
The agents are not aware of such a global labeling, but can differentiate between their ``neighbors''.
Neighbor relations among the $N$ agents are described by a graph $\bbb G=(\scr V,\scr E)$, called the neighbor graph, with vertex set $\scr V$ corresponding to agents and edge set $\scr E$ depicting neighbor relations. Specifically, agent~$j$ is a neighbor of agent $i$ whenever $(j,i)$ is an edge (or an arc) in $\bbb G$, representing that agent $i$ can receive information from agent $j$. For simplicity, we assume each agent $i$ is always a neighbor of itself, and thus $\bbb G$ has a self-arc at each vertex. We will consider both undirected and directed graphs in this paper. In undirected graphs, all edges are bi-directional, so if $i$ is $j$'s neighbor, $j$ is also $i$'s neighbor. 

All $N$ agents face a common set of $M$ arms (or decisions) which is denoted by $[M] \triangleq \{1,2,\ldots,M\}$. At each discrete time $t\in\{0,1,2,\ldots,T\}$, each agent $i$ makes a decision on which arm to select from the $M$ choices, and the selected arm is denoted by $a_i(t)$. If agent $i$ selects an arm $k$, it will receive a random reward $X_{i,k}(t)$. For each $i\in[N]$ and $k\in[M]$, $\{X_{i,k}(t)\}_{t=1}^T$ is an unknown i.i.d. random process. For each arm $k\in[M]$, all $X_{i,k}(t)$, $i\in[N]$, share the same expectation $\mu_{k}$. Without loss of generality, we assume that all $X_{i,k}(t)$ have bounded support $[0,1]$ and that $\mu_{1}\ge \mu_{2}\ge \cdots \ge\mu_{M}$, which implies that arm 1 has the largest reward mean and thus is always an optimal choice.

The goal of the decentralized multi-armed bandit problem just described is to devise a decentralized algorithm for each agent in the network which will enable agent $i$ to minimize its expected cumulative regret, defined as
\begin{align}
    R_{i}(T)&=T\mu_{1}-\sum_{t=1}^{T}\mathbf{E}\left[X_{a_{i}(t)}\right],\label{eq:regretdef1}
\end{align}
at an order at least as good as $R_i(T) = o(T)$, i.e., $R_i(T)/T \rightarrow 0$ as $T\rightarrow\infty$, for all $i\in[N]$.
More importantly, we require that each agent in the multi-agent network has improved, lower regret compared to that using the single-agent algorithm whenever it has at least one incoming neighbor in the network, which incentivizes collaboration among neighboring agents.

\section{Algorithms}\label{sec:algorithm}
We begin with some important variables to help present our algorithms. 

\textbf{Sample counter and local sampling estimate:}
Let $n_{i,k}(t)$ be the number of times
agent $i$ has pulled arm $k$ by time $t$.
Let $m_{i,k}(t)$ be agent $i$'s {\em estimate} of the maximal sampling times of arm $k$ until time $t$ over its neighborhood, which is updated as follows:
\begin{align}
    m_{i,k}(t+1)=& \;\max \Big\{n_{i,k}(t+1),\;\max_{j\in\scr N_i}m_{j,k}(t)\Big\}, \label{tilden}
\end{align}
where $\scr N_i$ stands for the neighbor set of agent $i.$
The variable $m_{i,k}(t)$ and its update \eqref{tilden} help agent~$i$ keep track of the maximal sampling times of arm $k$ among all those agents in the network which lie in the same connected component.

\textbf{Sample mean and reward mean estimate:}
Let $ \bar{x}_{i,k}(t)$ be the sample mean, representing the average reward that agent $i$ receives from arm $k$ until time $t$, which is updated as follows: 
\begin{align}\label{tildex}
    \bar{x}_{i,k}(t) = \frac{1}{n_{i,k}(t)}\sum_{\tau=0}^{t}\mathds 1(a_i(\tau)=k)X_{i,k}(\tau),
\end{align}
where $\mathds 1(\cdot)$ is the indicator function that returns 1 if the statement is true and 0 otherwise. 
Let $z_{i,k}(t)$ be agent $i$'s {\em estimate} of the reward mean of arm $k$ until time $t$, which is updated as follows:
\begin{align}
    &z_{i,k}(t+1)= \sum_{j\in\scr{N}_i}w_{ij}z_{j,k}(t)  +\bar{x}_{i,k}(t+1)-\bar{x}_{i,k}(t), \label{update}
\end{align}
where $w_{ij}$ are the entries of  an $N\times N$ matrix $W$. In the case when $\bbb G$ is an undirected graph, $W$ is a Metropolis matrix~\cite{xiao2005scheme} whose entries are defined as
\begin{align}
    w_{ij} &= \frac{1}{\max\{|\scr N_i|,|\scr N_j|\}}, \;\;\; j\in\scr N_i\setminus\{i\}, \;\;\; w_{ij} = 0, \;\;\; j\notin\scr N_i, \nonumber\\
    w_{ii} &= 1-\sum_{j\in\scr N_i}\frac{1}{\max\{|\scr N_i|,|\scr N_j|\}}.\label{eq:metroweights}
\end{align}
Here $|\scr N_i|$ stands for the number of neighbors of agent $i$.
It is worth emphasizing that each agent $i$ can obtain $|\scr N_j|$, $j\in\scr N_i$ from its neighbors through a one-time transmission, ensuring that the algorithm remains fully decentralized.
In the case when $\bbb G$ is a directed graph, $W$ is a flocking matrix~\cite{flocking} whose entries are defined as
\begin{align}
    w_{ij} &= \frac{1}{|\scr N_i|}, \;\;\; j\in\scr N_i, \;\;\; w_{ij} = 0, \;\;\; j\notin\scr N_i. \label{eq:flockweights}
\end{align}
It is worth emphasizing that a Metropolis matrix is always doubly stochastic, whereas a flocking matrix is always stochastic but may not be doubly stochastic. 
Let $\rho_2$ be the second largest magnitude among all the eigenvalues of $W$. It is well-known that if graph $\bbb G$ is connected  or strongly connected, there holds $\rho_2 < 1$.
The update \eqref{update} is thus a summation of an average consensus item $\sum_{j\in\scr{N}_i}w_{ij}z_{j,k}(t)$ and the term $\bar{x}_{i,k}(t+1)-\bar{x}_{i,k}(t)$ which can be regarded as a ``coarse gradient''. It is worth noting that the update \eqref{update} requires the reward at time $t+1$.
Let $z_{k}(t)$ and $\bar{x}_{k}(t)$ be the column stacks of all $z_{i,k}(t)$ and $\bar{x}_{i,k}(t)$, respectively. Then, the $N$ equations in \rep{update} can be combined~as 
\begin{align}
     z_{k}(t+1)=W z_{k}(t)+\bar{x}_{k}(t+1)-\bar{x}_{k}(t).\label{update:vector}
\end{align}

\textbf{Local arm index set:} Each agent $i$ keeps and updates an arm index set $\scr A_i(t)$ at each time $t$, which serves as the index collection of those arms that ``fall behind'' in exploring (i.e., $n_{i,k}(t)\leq m_{i,k}(t)-M$).


{\bf Local upper confidence bound function:}
Each agent $i$ needs to specify a design object in its local implementation, namely its upper confidence bound function, $Q_i(t)$ or $C(t, n_{i,k}(t))$, which will be used to quantify agent $i$'s belief on its estimate of arm $k$'s reward mean.
Upper confidence bound functions are critical in single-agent UCB algorithm design. As we will see, coordination among the agents allows us to design upper confidence bound functions ``better'' than those in the conventional single-agent UCB algorithms. Detailed expressions of $Q_i(t)$ and $C(t, n_{i,k}(t))$ will be specified in the theorems.

\textbf{Divergence:} Two kinds of divergence (distance) are used in this paper: the {\em Euclidean distance} and the {\em Bernoulli KL-divergence}. 

The Euclidean distance of scalar $p$ and $q$ is defined as $|p-q|.$ Specifically, we denote $\Delta_k = \mu_1-\mu_k$ as the Euclidean distance between $\mu_1$ and $\mu_k.$ Then, \eqref{eq:regretdef1} can also be expressed~as 
\begin{align}
    R_{i}(T)&=\sum_{k:\;\Delta_k>0} \mathbf{E}(n_{i,k}(t))\cdot\Delta_k\label{eq:regretdef2}.
\end{align}
The KL-divergence is defined as  \[d(p;q)=p\log\frac{p}{q}+(1-p)\log\frac{1-p}{1-q},\;\;\;p,q\in[0,1].\]
It is easy to verify that $d(p;q)$ is continuous and increasing for $q\in[p,1]$ and continuous and decreasing for $p\in[0,q].$ It also holds that $d(p;p)=0$ and $d(p;1)=\infty.$

\subsection{Decentralized KL-UCB}
We first look into the decentralized version of KL-UCB.
A detailed description of our algorithm is given as follows.

\textbf{Initialization:} At time $t=0$, each agent $i$ samples each arm $k$ exactly once, and sets $n_{i,k}(0)=1$, $z_{i,k}(0)=\bar{x}_{i,k}(0)=X_{i,k}(0)$, $m_{i,k}(0)=1$. 

Between clock times $t$ and $t+1$, $t\in\{0,1,\ldots,T\}$, 
each agent $i$ performs the steps enumerated below in the order indicated.



\begin{enumerate}
    \item \textbf{Decision Making:}
    Set the local arm index set as $\scr A_i(t) = \{k\in[M] : n_{i,k}(t)\leq m_{i,k}(t)-M\}$.
    \begin{enumerate}
    \item If $\scr A_i(t)$ is empty, agent $i$ selects an arm $a_i(t+1)$ that returns 
    \[\max\big\{q:n_{i,k}(t)d(z_{i,k}(t);q)\le Q_i(t)\big\},\] with ties broken arbitrarily, and receives reward $X_{i,a_i(t+1)}(t+1)$. 
    \item If $\scr A_i(t)$ is nonempty, agent $i$ randomly pulls one arm in $\scr A_i(t)$.
\end{enumerate}
\item \textbf{Transmission:} Agent $i$ broadcasts $m_{i,k}(t)$ and $z_{i,k}(t)$ for all $k\in[M]$; at the same time, agent $i$ receives $m_{j,k}(t)$ and $z_{j,k}(t)$ for all $k\in[M]$ from each of its neighbors $j\in \scr{N}_i$. 
\item \textbf{Updating:} Agent $i$ updates its variables by setting
    \begin{align*}
    n_{i,k}(t+1) &= \begin{cases}
    n_{i,k}(t)+1 & \mbox{if } k = a_i(t+1), \\
    n_{i,k}(t) & \mbox{if } k\neq a_i(t+1), 
    \end{cases}\\
    \bar{x}_{i,k}(t+1) &= \frac{1}{n_{i,k}(t+1)}\sum_{\tau=0}^{t+1}\mathds 1(a_i(\tau)=k)X_{i,k}(\tau),\\
    z_{i,k}(t+1) &= \sum_{j\in\scr{N}_i}w_{ij}(t)z_{j,k}(t) +\bar{x}_{i,k}(t+1) - \bar{x}_{i,k}(t), 
    \\
    m_{i,k}(t+1) &= \max \Big\{n_{i,k}(t+1),\;\max_{j\in\scr N_i}m_{j,k}(t)\Big\}.
    \end{align*}
\end{enumerate}


{\bf Key Design Ideas:}
The key difference of the decision making step compared with the single-agent KL-algorithm~\cite{KLUCB} is that each agent controls a set $\scr A_i(t)$ at each time instance to decide which rule to follow before choosing the arm. The bifurcated rules in our decision making step arise from the nature of decentralized cooperative algorithms: if one agent has poor estimation on an arm due to insufficient exploration, such poor estimation would keep getting amplified via information fusion over the network, and consequently becomes a hindrance on the estimation quality of all the agents. In other words, even if an agent itself has made sufficient exploration on one arm, it may still be ``misled'' by another agent if the latter has not, which may dramatically influence its estimation accuracy and the convergence rate; this is a critical feature and challenge in the decentralized setting. With this in mind, case (b) is thus designed to tackle the challenge by restricting a quantitative relation between agent $i$'s local sample count $n_{i,k}(t)$ and the estimated global maximum sampling times $m_{i,k}(t),$ which guarantees none of the agents ``falls behind'' too much (i.e., $n_{i,k}(t)\leq m_{i,k}(t)-M$) in the exploration process. While case (a), similar to the decision making step in the single-agent algorithm~\cite{KLUCB}, can be understood as each agent either chooses the arms that are not sufficiently explored (those with small $n_{i,k}(t)$), or the arms with large reward mean estimate (those with large $z_{i,k}(t)$), which forces agents to balance between exploration and exploitation, a major challenge of MAB problems. Together, both cases ensure that all agents will sufficiently explore each arm and that such exploration processes will always be kept ``on the same page'' for all~agents.



For a concise presentation of the algorithm, we refer to Algorithm~\ref{algorithm:KL} in the appendix. 



\subsubsection{Results}

\begin{theorem}\label{thm:KL}
Suppose that $\bbb G$ is undirected and connected, and that all $N$ agents adhere to Algorithm~\ref{algorithm:KL}. Then, with bounded rewards over $[0,1]$, Metropolis weights $w_{ij}$ given in \eqref{eq:metroweights}, and 
\[Q_i(t) = (1+\varsigma_i)\delta_i=3(1+\varsigma_i)(\log t+3\log\log t)/(2|\scr N_i|),\]
where $\varsigma_i$ is an arbitrary positive constant, the regret of each agent $i$ until time $T$ satisfies 
\begin{align*}
    R_i(T)\le \sum_{k:\Delta_k>0} (K_{i,k}(T) + \Psi(\epsilon, T))\Delta_k,
\end{align*}
where $\Psi(\epsilon, T)$ is a positive function satisfying $\Psi(\epsilon, T)=O(\log\log T),$ and 
\[K_{i,k}(T)=\max\bigg\{2F_2(\kappa_i),\bigg\lfloor\frac{(1+\varsigma_i)(1+\epsilon)\delta_i}{d(\mu_k;\mu_1)}\bigg\rfloor\bigg\},\]
with $\epsilon$ being a positive constant, $F_2(\cdot)$ defined in Remark~\ref{re:def}, and $\kappa_i$ satisfying \eqref{eq:tildevarsigma_i}.
\end{theorem}

It is worth mentioning that $\kappa_i$, which satisfies \eqref{eq:tildevarsigma_i}, is unique. The proof of this uniqueness can be found immediately following Eq. \eqref{eq:tildevarsigma_i} in the proof of Lemma \ref{le:KLpart1}.



\vspace{.2in}

\begin{remark}\label{re:def}
Define a function $F_2:\R\rightarrow\R$ by $F_2(\epsilon) = \max\{f(\epsilon), 2(M^2+2MN+N)\}$, where $f(\epsilon)$ is the smallest value such that, when a positive integer $n\ge f(\epsilon)$, there holds
\begin{align*}
    \bigg(\rho_2^{n}+\sum_{h=2}^{n}\frac{\rho_2^{n-h}}{(h-1)h}\bigg)n^{\frac{3}{2}}\le\frac{\epsilon[W_\infty]_{ij}}{\sigma} \;\;\; {{\rm and}} \;\;\;
    \frac{2}{n}\log_{1/\rho_2}n<1.
\end{align*}
Here $\sigma$ is defined in Lemma~\ref{le:directrho2}, and $W_\infty\triangleq\lim_{t\rightarrow\infty}W^t$, whose existence is guaranteed by Lemma~\ref{le:directrho2}.  It is clear that $W_\infty=\frac{1}{N}\1\1^\top$ if in addition $W$ is doubly stochastic.
It will be shown respectively in the proofs of Lemma~\ref{le:distance} and Lemma~\ref{le:KLpart1} that $F_2(\cdot)$ and $\kappa_i$ are well-defined. 
\hfill$\Box$
\end{remark}

\vspace{.2in}

\begin{remark}\label{re:outperformKL}
\normalfont
From Theorem~\ref{thm:KL}, the asymptotic upper bound of the regret for each agent $i$ satisfies
\begin{align*}
    \lim_{T\rightarrow\infty}\frac{R_i(T)}{\log T}\le \sum_{k:\Delta_k>0}\frac{3(1+\varsigma_i)}{2|\scr N_i|d(\mu_k;\mu_1)}.
\end{align*}
First, it implies that each agent $i$'s regret bound asymptotically reciprocal to its neighbor set size. Since each agent $i$ is always assumed to be its own neighbor, the maximum possible asymptotic regret bound occurs when $|\scr N_i| = 2$. Second,
it is easy to see that the lower $\varsigma_i$ is set, the lower the asymptotic upper bound would be, and this pattern is reflected in the finite-time simulations; see Figure~\ref{fig:dist_KL_vs_single_KL}, Figure~\ref{fig:dist_KL_vs_dist_UCB1} and Section~\ref{appendix:simulation}. Thus, in practice, we recommend choosing a low $\varsigma_i$ value for each agent, e.g., let $Q_i(t) = 3.1(\log t+3\log\log t)/(2|\scr N_i|).$ 
It is worth emphasizing that the above  asymptotic regret bound also reflects the local network effect, that is, the more neighbors an agent has, the smaller regret bound it achieves. 
For the case when each agent $i$ independently uses the single-agent KL-UCB algorithm in \cite{KLUCB}, the corresponding regret~satisfies
\begin{align*}
    \lim_{T\rightarrow\infty}\frac{R_i(T)}{\log T}\le \sum_{k:\Delta_k>0}\frac{1}{d(\mu_k;\mu_1)}.
\end{align*}
As long as each $\varsigma_i$ is set to be smaller than $2|\scr N_i|/3 - 1$ (or $1/3$ considering the worst case when $|\scr N_i|=2$), the asymptotic regret bound of our decentralized algorithm is guaranteed to be better than the single-agent counterpart.
\hfill$\Box$
\end{remark}

Since the reward distribution is homogeneous, each agent can learn the reward mean without information from others, let alone information from the whole graph. With this in mind, we now remove the connectivity requirement of $\bbb G.$ From the fact that $\bbb G$ is undirected, it can be divided into a set of connected components. Let $\bbb G_i$ be the largest connected component that contains agent $i$. Then, Theorem~\ref{thm:KL} is applicable to $\bbb G_i.$ Combining the result with Remark~\ref{re:outperformKL}, we obtain the following~corollary.

\vspace{.2in}

\begin{corollary}\label{co:requirenoconnectness}
Suppose that $\bbb G$ is an arbitrary undirected graph, and that each agent adheres to Algorithm~\ref{algorithm:KL} when it is not isolated\footnote{An agent is isolated if it has no neighbor apart from itself.} and otherwise adheres to the single-agent KL-UCB. Then, with bounded rewards over $[0,1]$, the same $Q_i(t)$ in Theorem~\ref{thm:KL},
and $\varsigma_i<1/3,$ the asymptotic regret bound of each agent is strictly lower than the single-agent counterpart whenever the agent is not isolated.
\end{corollary}

\vspace{.2in}

\begin{remark}\label{re:knowN}
From Theorem~\ref{thm:KL}, if $\bbb G$ is a complete graph in which $|\scr N_i|=N$ for all $i$, the asymptotic regret bound for each agent becomes $
    O(\log T/N),
$
which is the largest ``collaborative gain'' in regret improvement that it could possibly be for an $N$-agent network. Such a largest collaborative gain can also be achieved if each agent is assumed to be aware of the network size $N$. In this case,  with
the same $Q_i(t)$ in Theorem~\ref{thm:KL}
and the same arguments in its proof, it can be shown that 
\begin{align*}
    \lim_{T\rightarrow\infty}\frac{R_i(T)}{\log T}\le \sum_{k:\Delta_k>0}\frac{3(1+\varsigma_i)}{2Nd(\mu_k;\mu_1)}.
\end{align*}
Similar arguments also hold for Theorem~\ref{thm:editedmaintheorem}.
\hfill$\Box$
\end{remark}

{\bf Lower Bound:} From \cite{KLUCB}, when the reward distribution is Bernoulli, the lower bound of regret for single-agent KL-UCB satisfies
\begin{align*}
    R_i(T)\ge \sum_{k:\;\Delta_k>0}\bigg(\frac{\Delta_k}{d(\mu_k;\mu_1)}+o(1)\bigg)\log T.
\end{align*}
Among all connected graphs with $N$ agents, a complete graph leads to the best network regret (i.e., the sum of all $N$ agents' regrets) because each agent can receive information from all the agents. This is essentially equivalent to a single-agent case in which the agent can pull $N$ times at each time. Thus, the lower bound of each agent in an $N$-agent graph satisfies
\begin{align*}
    R_i(T)\ge \sum_{k:\;\Delta_k>0}\bigg(\frac{\Delta_k}{Nd(\mu_k;\mu_1)}+o(1)\bigg)\log T.
\end{align*}
Compared with the result in Remark~\ref{re:knowN}, when agents are aware of $N$ or when the graph is complete, the asymptotic regret bound of our algorithm is ``almost'' optimal for Bernoulli reward distributions.

For general distributions, the lower bound for single-agent algorithm is $\Omega(\log T)$~\cite{LR85}, and thus with the same claim above, the lower bound of each agent in an $N$-agent graph is $\Omega(\log T/N).$


{\bf Proof Sketch of Theorem~\ref{thm:KL}:}
As ``Key Design Ideas'' have pointed out, the key technical challenges for MAB problems and for decentralized algorithms are resolved respectively in the two cases of the decision making step of Algorithm~\ref{algorithm:KL}, and our proof is also centered around the two aspects. Specifically, Lemmas~\ref{le:directrho2} and \ref{le:Hoff} discuss some basic properties; Lemma~\ref{relation2} illustrates that our algorithm ensures the exploration consistency among all the agents; Lemmas~\ref{le:KLresult} and \ref{le:KLpart2} are the multi-agent decentralized extension of the single-agent KL-UCB results in \cite{KLUCB} and discuss the function of case (a); Lemmas~\ref{le:distance} and \ref{le:KLpart1} are the critical results which combine our decentralized algorithm design with both decision making cases, and imply that both the Euclidean distance and KL-divergence of the actual reward mean $\mu_k$ and each agent's estimated reward mean $z_{i,k}(t)$ converge to 0 in probability after a certain amount of samples. This is explained in the proof of Theorem~\ref{thm:KL} in detail. With these results in hand, we divide the proof of Theorem~\ref{thm:KL} into three parts. The first two parts apply the logic of analysis in single-agent KL-UCB~\cite{KLUCB} to our decentralized algorithm design, and together return each agent's expected sample counts on each arm due to case (a) after a certain amount of samples. The last part makes use of the design of $m_{i,k}(t)$ and returns each agent's expected number of samples on each arm due to case (b) after a certain amount of samples. The regret is thus at hand. Detailed proofs can be found in Section~\ref{sec:analysis}.
\hfill$\qed$

We further consider a scenario in which different agents start at different times, resulting in only a subset of agents being initialized at time 0. Let $D$ represent the first time instance when all agents become active. Compared to Algorithm 1, the only difference lies in the initial values of the variables for each agent. 
As will be seen in Section~\ref{sec:analysis}, the analysis of Algorithm 1 related to the arm reward mean estimates $z_{i,k}(t)$ is independent of the initial values. Thus, we only need to analyze the sample counter $n_{i,k}(t)$ and the local sampling estimates $m_{i,k}(t)$ for each agent in the scenario under consideration.
As mentioned in the Proof Sketch of Theorem 1, the local arm index set $\scr A_i(t)$ in Algorithm 1 is designed to guarantee and characterize exploration consistency among all agents, as stated in Lemma \ref{relation2}. To ensure the same exploration consistency for the scenario where agents start at different times, we only need to redefine the local arm index set as $\scr A_i(t) = \{k\in[M] : n_{i,k}(t)\leq m_{i,k}(t)-M-D\}$. 
With the new definition of $\scr A_i(t)$, the entire analysis process remains unchanged. 

Using the same arguments as in the analyses of Algorithm 1, including Lemmas 1--7 and the Proof of Theorem 1, with the corresponding adjustments for the steps following Eq. (13) in the supplementary material, the following result can be straightforwardly obtained. 

\vspace{.2in}

\begin{corollary}
Let $D$ be the first time instance at which all $N$ agents become active. With Algorithm~\ref{thm:KL} modified so that $\scr A_i(t)$ is replaced by $\{k\in[M] : n_{i,k}(t)\leq m_{i,k}(t)-M-D\}$, and under the same conditions as in Theorem 1, the regret of each agent $i$ until time $T$ satisfies
\begin{align*}
    R_i(T)\le \sum_{k:\Delta_k>0} (K'_{i,k}(T) + \Psi'(\epsilon, T))\Delta_k,
\end{align*}
where $\Psi'(\epsilon, T)=\Psi(\epsilon,T) + MD =O(\log\log T)$ and 
$$K'_{i,k}(T)=\max\bigg\{2F'_2(\kappa_i),\bigg\lfloor\frac{(1+\varsigma_i)(1+\epsilon)\delta_i}{d(\mu_k;\mu_1)}\bigg\rfloor\bigg\},$$
with $F'_2(\kappa_i) = \max\{f(\kappa_i), 2(M^2+2MN+MD+N)\}$, $\Psi(\epsilon,T)$ and $\delta_i$ defined in Theorem \ref{thm:KL}, and $f(\cdot)$ defined in Remark~\ref{re:def}.
\end{corollary}

Compared to Theorem \ref{thm:KL}, the corollary implies that the regret upper bound for each agent may increase by at most a constant if the multi-agent network takes $D$ time steps for all agents to become active. This constant is proportional to the value of $D$.

Similar results can be derived for the decentralized UCB1 algorithm in the next subsection.

\subsection{Decentralized UCB1}

In this section, we study the decentralized version of classic UCB1~\cite{auer2002finite}. The decentralized UCB1 shares the same structure of the decentralized KL-UCB introduced above, the only difference is case (a) of the decision making step, which is modified as:
\begin{enumerate}
    \item[(a)] If $\scr A_i(t)$ is empty, agent $i$ selects an arm $a_i(t+1)$ which returns largest \[z_{i,k}(t) + C(t, n_{i,k}(t)),\] with ties broken arbitrarily, and receives reward $X_{i,a_i(t+1)}(t+1)$.
\end{enumerate}
Here $C(t, n_{i,k}(t))$ is known as the upper confidence bound, which is designed to be a decreasing function of $n_{i,k}(t)$ and will be specified in the following theorem statement.

For a concise presentation of the algorithm, we refer to Algorithm~\ref{algorithm:UCB1} in the appendix.

It is worth emphasizing that the design of the decision-making bifurcation for an empty $\scr A_i(t)$ is not the only difference between Algorithm~\ref{algorithm:KL} and Algorithm~\ref{algorithm:UCB1}. Another, more important difference is the design of the UCB function, which is respectively given in the statements of the theorems.

\subsubsection{Results}

\begin{theorem}\label{thm:editedmaintheorem}
Suppose that $\bbb G$ is undirected and connected, and that all $N$ agents adhere to Algorithm~\ref{algorithm:UCB1}. Then, with bounded rewards over $[0,1]$, Metropolis weights $w_{ij}$ given in \eqref{eq:metroweights}, and 
\eq{C(t,n_{i,k}(t))=(1+\beta_i)\sqrt{\frac{3\log t}{|\scr N_i|n_{i,k}(t)}},\label{eq:Cfunction}}
where $\beta_i$, $i\in[N]$ are arbitrary positive constants, the regret of each agent $i$ until time $T$ satisfies 
\begin{align*}
R_i(T)\le \sum_{k:\Delta_k>0}\Big(\max\Big\{\frac{12(1+\beta_i)^2\log T}{|\scr N_i|\Delta_{k}^2},2F_2(\beta_i)\Big\}+\Gamma\Big)\Delta_k,
\end{align*}
where $F_2(\cdot)$ is defined in Remark~\ref{re:def}, and $\Gamma$ is defined as
\begin{align*}
    \Gamma &= M^2+2MN+N + \sum_{i\in[N]}\bigg(\frac{\pi^2}{3}+2F_2(\beta_i)-1\bigg).
\end{align*} 
\end{theorem}


\begin{remark}\label{re:outperformUCB1}
From Theorem~\ref{thm:editedmaintheorem}, the asymptotic bound of the regret for each agent $i$ satisfies
\begin{align*}
    \lim_{T\rightarrow\infty}\frac{R_i(T)}{\log T}\le \sum_{k:\Delta_k>0}\frac{12(1+\beta_i)^2}{|\scr N_i|\Delta_{k}}.
\end{align*}
It is easy to see a lower $\beta_i$ corresponds to a lower asymptotic regret, and generally this pattern also holds in finite-time simulations; see Figure~\ref{fig:dist_UCB1_vs_single_UCB1}, Figure~\ref{fig:dist_KL_vs_dist_UCB1}, and Section~\ref{appendix:simulation}. In practice, we can always set each $\beta_i$ to be a small value, e.g., $C(t,n_{i,k}(t)) = 1.01\sqrt{\frac{3\log t}{|\scr N_i|n_{i,k}(t)}}.$
For the case when each agent $i$ independently uses the single-agent UCB1 algorithm in \cite{auer2002finite}, the regret satisfies
\begin{align*}
    \lim_{T\rightarrow\infty}\frac{R_i(T)}{\log T}\le \sum_{k:\Delta_k>0}\frac{8}{\Delta_{k}}.
\end{align*}
Thus, as long as each $\beta_i$ is set to be smaller than $\sqrt{2|\scr N_i|/3} - 1$ (or $\sqrt{\frac{4}{3}} - 1$ considering the worst case when $|\scr N_i| = 2$), the asymptotic regret bound of our decentralized algorithm is ensured to be better than the single-agent counterpart.
\hfill$\Box$
\end{remark}

Similar to Corollary~\ref{co:requirenoconnectness}, we have the following result.

\vspace{.2in}

\begin{corollary}
Suppose that $\bbb G$ is an arbitrary undirected graph, and that each agent adheres to Algorithm~\ref{algorithm:UCB1} when it is not isolated and otherwise adheres to the single-agent UCB1. Then, with bounded rewards over $[0,1],$  the same  $C(t,n_{i,k}(t))$ in Theorem~\ref{thm:editedmaintheorem},
 and $\beta_i< \sqrt{\frac{4}{3}} - 1,$ the asymptotic bound of each agent better than the single-agent counterpart whenever it is not isolated.
\end{corollary}

Detailed clarification and techniques can be found in 
the next section.
The following theorem further extends our algorithm design framework to encompass directed graphs.


\vspace{.2in}

\begin{theorem}\label{thm:directed}
    Suppose that $\bbb G$ is directed and strongly connected, and that all $N$ agents adhere to Algorithm~\ref{algorithm:UCB1}. Then, with bounded rewards over $[0,1]$, flocking weights $w_{ij}$ given in \eqref{eq:flockweights}, and 
\[C(t,n_{i,k}(t))=(1+\beta_i)\sqrt{\frac{3\log t}{2n_{i,k}(t)}},\]
where $\beta_i$, $i\in[N]$ are arbitrary positive constants, the regret of each agent $i$ until time $T$ satisfies 
\begin{align*}
&\quad\;R_i(T)\\&\le \sum_{k:\Delta_k>0}\Big(\max\Big\{\frac{6(1+\beta_i)^2\log T}{\Delta_{k}^2},2F_2(\beta_i)\Big\}+\Gamma\Big)\Delta_k,
\end{align*}
where $F_2(\cdot)$ and $\Gamma$ are defined in Remark~\ref{re:def}. 
\end{theorem}


{\bf Proof Sketch of Theorems~\ref{thm:editedmaintheorem} and \ref{thm:directed}:} 
The primary objective of the analysis is to address equation \eqref{eq:totalestimate}. To evaluate the first summation term in our decentralized scenario, we utilize two key properties: exploration consistency, as illustrated in Lemma~\ref{relation2}, and the diminishing distance between $z_{i,k}(t)$ and $\mu_k$, as demonstrated in Lemma~\ref{le:distance}. Notably, Lemma~\ref{le:distance} is a generalized property applicable to both directed and undirected graphs. The estimation results for the aforementioned probability are provided in \eqref{eq:undirected_1/t^4} in the proof of Theorem~\ref{thm:editedmaintheorem} and \eqref{eq:directed_1/t^4} in the proof of Theorem~\ref{thm:directed}. The evaluation of the second summation term in \eqref{eq:totalestimate} is detailed in Part B in the proof of Theorem~\ref{thm:editedmaintheorem}, primarily leveraging the exploration consistency discussed in Lemma~\ref{relation2}.
\hfill$\qed$

The existing fully decentralized algorithm in \cite{jingxuan2021} only works for undirected graphs. 
Another fully decentralized algorithm may be possibly constructed by applying the heterogeneous decentralized algorithm in \cite{tac_mab} to the homogeneous case. However, it relies on a doubly stochastic consensus update matrix and thus implicitly requires that the underlying graph be undirected or balanced\footnote{A weighted directed graph is called balanced if the sum of all in-weights equals the sum of all out-weights at each of its vertices \cite{gharesifard2013distributed}.}. It is unclear how its algorithm design can be applied to general unbalanced directed graphs. The algorithm in \cite{acc21}, to our knowledge, is the only one in the existing literature crafted for general directed graphs, but it requires each agent be aware of the network size $N$.

\vspace{.2in}

\begin{remark}\label{re:outperformUCB1directed}
From Theorem~\ref{thm:directed}, the asymptotic bound of the regret for each agent $i$ satisfies
\begin{align*}
    \lim_{T\rightarrow\infty}\frac{R_i(T)}{\log T}\le \sum_{k:\Delta_k>0}\frac{6(1+\beta_i)^2}{\Delta_{k}}.
\end{align*}
It is easy to see a lower $\beta_i$ corresponds to a lower asymptotic regret, and generally this pattern also holds in finite-time simulations; see Fig.~\ref{fig:directed_normal} and Fig~\ref{fig:directed_beta}. 
In practice, we can always set each $\beta_i$ to be a small value, e.g., $C(t,n_{i,k}(t)) = 1.01\sqrt{\frac{3\log t}{2n_{i,k}(t)}}.$
For the case when each agent $i$ independently uses the single-agent UCB1 algorithm in \cite{auer2002finite}, the regret satisfies
$
    \lim_{T\rightarrow\infty}\frac{R_i(T)}{\log T}\le \sum_{k:\Delta_k>0}\frac{8}{\Delta_{k}}
$.
Thus, as long as each $\beta_i$ is set to be smaller than $\frac{2}{\sqrt{3}} - 1$, each agent's asymptotic regret bound of our decentralized UCB1 algorithm is ensured to be better than the single-agent counterpart, for any directed, possibly unbalanced, strongly connected neighbor graphs.
\hfill$\Box$
\end{remark}

It is worth noting that Theorem \ref{thm:directed} only guarantees a constant coefficient improvement in asymptotic regret and thus does not reflect any
local neighborhood effect. It turns out that how to fully leverage neighbor information/collaboration in decentralized UCB1 algorithm design is challenging for general directed graphs. Specifically, if one still chooses the same local upper confidence bound function \eqref{eq:Cfunction}, which was designed for undirected graphs and embeds the local incoming neighbor set size, for directed graphs, the results in \eqref{eq:directed_1/t^4} will no longer hold. 
How to tackle this challenging problem has so far eluded us and is a direction for future work.

\section{Analysis}\label{sec:analysis}

In this section, we provide the analysis of Algorithms~\ref{algorithm:KL} and \ref{algorithm:UCB1}, as well as the proofs of Theorems \ref{thm:KL} and \ref{thm:editedmaintheorem}. The proofs of all lemmas are omitted due to space limitations and can be found in the supplementary material version which will be uploaded to arXiv after acceptance.

\subsection*{Preliminary results:}





\begin{lemma}\label{le:directrho2}
    If $W$ is an irreducible stochastic matrix with positive diagonal entries, then for any $i,j\in[N]$, there exists a positive vector $a$ such that $\1'a=1$ and $W_\infty=\1a'$. 
Moreover, there exists a positive $\sigma$ for which
    \begin{align*}
        \|W^t-W_\infty\|\le \sigma\rho_2^t,\;\;\;\;\; |[W^t]_{ij}-[W_\infty]_{ij}|\le \sigma\rho_2^t.
    \end{align*}
\end{lemma}



The proof of Lemma~\ref{le:directrho2} is standard, and can be found in, say \cite[Theorem 4.9]{levin2017markov}.

\begin{lemma}\label{le:Hoff}
{\rm (Hoeffding's inequality \cite{Hoeffding1963})} 
Let $\{X_1,\ldots,X_n\}$ be a finite set of independent random variables such that $\mathbf{E}(X_i) = \mu_i$ and $X_i\in[a_i, b_i]$ for all $i\in [n]$. Then, for any $\eta\ge 0,$ 
\begin{equation}\label{Hoeffding2}
    \begin{split}
        \mathbf{P}(X_1+\cdots+X_n-(\mu_1+\cdots+\mu_n)\ge\eta)
        &\le\exp\bigg(\frac{-2\eta^2}{\sum_{i=1}^n(b_i-a_i)^2}\bigg),\\
    \mathbf{P}(X_1+\cdots+X_n-(\mu_1+\cdots+\mu_n)\le-\eta)
    &\le\exp\bigg(\frac{-2\eta^2}{\sum_{i=1}^n(b_i-a_i)^2}\bigg).
    \end{split}
\end{equation}
\end{lemma}

\subsection*{Exploration consistency:}








\begin{lemma}\label{relation2}
{\rm (\!\!\cite[Lemma 6]{tac_mab})}
For any $i,h\in[N]$ and $k\in[M]$,
it holds that 
\begin{align*}
    |n_{i,k}(t)-n_{h,k}(t)|\le M^2+2MN+N.
\end{align*}
Consequently, if $n_{i,k}(t)\geq 2(M^2+2MN+N)$, then for any $h\in[N],$  
\[\frac{1}{2}n_{h,k}(t)\le n_{i,k}(t)\le \frac{3}{2}n_{h,k}(t).\]
\end{lemma}

\subsection*{Preparation for bounding the Euclidean distance between $\mu_k$ and $z_{i,k}(t)$:}








\begin{lemma}\label{le:distance}
    For any $i\in[N]$ and $k\in[M],$ when $n_{i,k}(t)\ge 2F_2(\epsilon),$ where $F_2(\epsilon)$ is defined in Remark~\ref{re:def}, it holds for any $\epsilon>0$ that
\begin{align*}
    z_{i,k}(t)\le (1+\epsilon)\sum_{j\in[N]}[W_\infty]_{ij}\bar x_{j,k}(t),\\
    z_{i,k}(t)\ge(1-\epsilon)\sum_{j\in[N]}[W_\infty]_{ij}\bar x_{j,k}(t).
\end{align*}
\end{lemma}

{\bf Proof of Lemma~\ref{le:distance}:}
Expanding \eqref{update:vector} leads to
\begin{align*}
   z_{k}(t)&= W\cdot z_k(t-1)+\bar{x}_k(t)-\bar{x}_k(t-1)\\
   &=W^t\cdot z_k(0)+\sum_{\tau=0}^{t-1}W^{\tau}(\bar{x}_k(t-\tau)-\bar{x}_k(t-\tau-1))\\
   &=\sum_{\tau=0}^{t-1}(W^{t-\tau}-W^{t-\tau-1})\bar{x}_k(\tau)+\bar{x}_k(t).
\end{align*}
Then, for each agent $i,$
\begin{align}\label{summationofz}
    z_{i,k}(t)=\sum_{j\in[N]}\bigg\{\sum_{\tau=0}^{t-1}[W^{t-\tau}-W^{t-\tau-1}]_{ij}\bar{x}_{j,k}(\tau)+[W^0]_{ij}\bar{x}_{j,k}(t)\bigg\}.
\end{align}
For each $j\in[N],$ let $\tau_{j,k,1},\ldots,\tau_{j,k,n_{j,k}(t)}$ denote the sequence of time instances such that $a_j(t) = k$ till time $t.$ Then, \eqref{summationofz} can be further written as 
\begin{align*}
    z_{i,k}(t)
    = \sum_{j\in[N]}\sum_{h=1}^{n_{j,k}(t)-1}[W^{t-\tau_{j,k,h}}-W^{t-\tau_{j,k,h+1}}]_{ij}\bar x_{j,k}(\tau_{j,k,h})
    +[W^{t-\tau_{j,k,n_{j,k}(t)}}]_{ij}\bar x_{j,k}(\tau_{j,k,n_{j,k}(t)}).
\end{align*}
From the above expression and the definition of $\bar x_{i,k}(t)$, it is easy to see that $z_{i,k}(t)$ is a linear combination of $X_{j,k}(t)$ for $j\in[N]$ and $t>0.$ Let $c_{i,j,k,n}(t)$ be the coefficient of $X_{j,k}(n)$ in the pre-mentioned linear combination.Then, when $a_j(n)\neq k,$ $c_{i,j,k,n} = 0,$ and when $a_j(n)= k,$ 
\begin{align*}
    c_{i,j,k,n}(t) = \bigg[\sum_{h=n}^{n_{j,k}(t)-1}\frac{W^{t-\tau_{j,k,h}}-W^{t-\tau_{j,k,h+1}}}{h} 
    +\frac{W^{t-\tau_{j,k,n_{j,k}(t)}}}{n_{j,k}(t)}\bigg]_{ij}
    =\bigg[\frac{W^{t-\tau_{j,k,n}}}{n}-\sum_{h=n+1}^{n_{j,k}(t)}\frac{W^{t-\tau_{j,k,h}}}{(h-1)h}\bigg]_{ij}.
\end{align*}
Since for any $n\le n_{j,k}(t),$ $\frac{1}{n}-\sum_{h=n+1}^{n_{j,k}(t)}\frac{1}{(h-1)h}=\frac{1}{n_{j,k}(t)},$ 
from Lemma~\ref{le:directrho2}, for any $n$ such that $a_{j}(n) = k,$ it holds that 
\begin{align}\label{eq:coefficient}
    \bigg|c_{i,j,k,n}(t)-\frac{[W_\infty]_{ij}}{n_{j,k}(t)}\bigg|
    &\le \frac{[W^{t-\tau_{j,k,n}}-W_\infty]_{ij}}{n}+\sum_{h=n+1}^{n_{j,k}(t)}\frac{[W^{t-\tau_{j,k,h}}-W_\infty]_{ij}}{(h-1)h}\nonumber\\
    &\le \frac{\sigma\rho_2^{t-\tau_{j,k,n}}}{n}+\sum_{h=n+1}^{n_{j,k}(t)}\frac{\sigma\rho_2^{t-\tau_{j,k,h}}}{(h-1)h}\nonumber\\
    &\le \sigma\rho_2^t+\sum_{h=2}^{n_{j,k}(t)}\frac{\sigma\rho_2^{t-\tau_{j,k,h}}}{(h-1)h}.
\end{align}
Since $0<\rho_2<1,$ the smaller $t-\tau_{j,h}$ is, the larger the right side of the inequality would be. Since $t-\tau_{j,h}\le n_{j,k}(t)-h$ by definition, we have $\rho_2^{t-\tau_{j,k,h}}\leq \rho_2^{n_{j,k}(t)-h}$. Thus, for any $\bar t\in(2, n_{j,k}(t)),$
\begin{align}
    \sum_{h=2}^{n_{j,k}(t)}\frac{\rho_2^{t-\tau_{j,k,h}}}{(h-1)h}
    &\le\sum_{h=2}^{n_{j,k}(t)}\frac{\rho_2^{n_{j,k}(t)-h}}{(h-1)h}\nonumber\\
    &= \bigg(\sum_{h=2}^{\bar t}+\sum_{h=\bar t+1}^{n_{j,k}(t)}\bigg)\frac{\rho_2^{n_{j,k}(t)-h}}{(h-1)h}\nonumber\\
    &= \sum_{h=1}^{\bar t-1}\frac{\rho_2^{n_{j,k}(t)-h-1}}{h} - \sum_{h=2}^{\bar t}\frac{\rho_2^{n_{j,k}(t)-h}}{h}+\sum_{h=\bar t+1}^{n_{j,k}(t)}\frac{\rho_2^{n_{j,k}(t)-h}}{(h-1)h}\nonumber\\
    &=\rho_2^{n_{j,k}(t)-2}+\sum_{h=2}^{\bar t-1}\frac{\rho_2^{n_{j,k}(t)-h-1}(1-\rho_2)}{h}-\frac{\rho_2^{n_{j,k}(t)-\bar t}}{\bar t}
    +\sum_{h=\bar t+1}^{n_{j,k}(t)}\frac{\rho_2^{n_{j,k}(t)-h}}{(h-1)h}\nonumber\\
    &\le \rho_2^{n_{j,k}(t)-2}+\sum_{h=2}^{\bar t-1}\rho_2^{n_{j,k}(t)-h-1}(1-\rho_2)+\sum_{h=\bar t+1}^{n_{j,k}(t)}\frac{1}{(h-1)h}\nonumber\\
    &=\rho_2^{n_{j,k}(t)-\bar t}+\frac{1}{\bar t}-\frac{1}{n_{j,k}(t)}.\nonumber
\end{align}
Combining the result with \eqref{eq:coefficient}, for any $n$ such that $a_{j}(n) = k,$ it holds that 
\begin{align*}
    \bigg|c_{i,j,k,n}(t)-\frac{[W_\infty]_{ij}}{n_{j,k}(t)}\bigg|\le 2\rho_2^{n_{j,k}(t)-\bar t}+\frac{1}{\bar t}-\frac{1}{n_{j,k}(t)}.
\end{align*}
Now letting $\bar t = n_{j,k}(t) - 2\log_{\frac{1}{\rho_2}} n_{j,k}(t),$ it holds that 
\begin{align*}
    2\rho_2^{n_{j,k}(t)-\bar t}+\frac{1}{\bar t}-\frac{1}{n_{j,k}(t)}
    = \frac{2}{n^2_{j,k}(t)} +\frac{1}{n_{j,k}(t) - 2\log_{\frac{1}{\rho_2}} n_{j,k}(t)}-\frac{1}{n_{j,k}(t)}
    = o(n^{-\frac{3}{2}}_{j,k}(t)).
\end{align*}
From Remark~\ref{re:def}, when $n_{j,k}(t)\ge f(\epsilon),$ it holds that 
\begin{align}\label{eq:coefficientbound}
    \bigg|c_{i,j,k,n}(t)-\frac{[W_\infty]_{ij}}{n_{j,k}(t)}\bigg|&\le \epsilon[W_\infty]_{ij}n^{-\frac{3}{2}}_{j,k}(t).
\end{align}
From Lemma~\ref{relation2}, it is easy to see that when $n_{i,k}(t)\ge 2F_2(\epsilon),$ it holds for all $j\in[N]$ that $n_{j,k}(t)\ge F_2(\epsilon)\ge f(\epsilon),$ then 
\begin{align*}
    &\quad\;\bigg|\sum_{j\in[N]}\sum_{n=1}^{n_{j,k}(t)}\bigg(c_{i,j,k,\tau_{j,k,n}}(t)-\frac{[W_\infty]_{ij}}{n_{j,k}(t)}\bigg)X_{j,k}(\tau_{j,k,n})\bigg|\\
    &\le \sum_{j\in[N]}\sum_{n=1}^{n_{j,k}(t)}\epsilon[W_\infty]_{ij}\sqrt{n_{j,k}(t)}\frac{X_{j,k}(\tau_{j,k,n})}{n_{j,k}(t)}\\
    &\le\sum_{j\in[N]}\sum_{n=1}^{n_{j,k}(t)}\epsilon[W_\infty]_{ij}\frac{X_{j,k}(\tau_{j,k,n})}{n_{j,k}(t)}.
\end{align*}
Substituting $\sum_{j\in[N]}\sum_{n=1}^{n_{j,k}(t)}c_{i,j,k,\tau_{j,k,n}}(t)X_{j,k}(\tau_{j,k,n})=z_{i,k}(t)$ and \eqref{tildex} to the above inequality, we obtain that
\begin{align*}
    \bigg|z_{i,k}(t)-\sum_{j\in[N]}[W_\infty]_{ij}\bar x_{j,k}(t)\bigg|\le \epsilon\sum_{j\in[N]}[W_\infty]_{ij}\bar x_{j,k}(t),
\end{align*}
which completes the proof.
\hfill$\qed$


\subsection*{Preparation for bounding the KL divergence between $\mu_k$ and $z_{i,k}(t)$:}

\begin{lemma}\label{le:KLresult}
Let $\delta_i$ be an arbitrary positive constant,  $t_m = \lfloor(1+\eta_i)^m\rfloor$ with $\eta_i = \frac{1}{\delta_i - 1}$ for $m>0$, $s$ be the smallest integer such that $\frac{\delta_i}{s+1}\le d(0;\mu_k),$ and $\tilde t_m = \max(s, t_m).$ Define $\phi_{\mu_k}(\lambda) = \log \mathbf E(\exp(\lambda X_{i,k}(0)))$ and $\lambda(x) = \log\frac{x(1-\mu_k)}{\mu_k(1-x)}$. If $l_m<\mu_k$ such that $d(l_m;\mu_k)=\frac{\delta_i}{(1 + \eta_i)^m}$,
then 
\begin{align*}
    &\quad\;\mathbf P\Big(\tilde t_{m-1}<n_{i,k}(t)\le t_m, \bar X_{i,k}(t)<\mu_k,
    d(\bar X_{i,k}(t);\mu_k)\ge\frac{\delta_i}{n_{i,k}(t)}\Big)\\
    &\le \mathbf P\Big(\lambda(l_m)\bar X_{i,k}(t)-\phi_{\mu_k}(\lambda(l_m))\ge \frac{\delta_i}{n_{i,k}(t)(1+\eta_i)}\Big)\\
    &\le\exp\Big(\frac{-\delta_i}{1+\eta_i}\Big).
\end{align*}
\end{lemma}


Lemma \ref{le:KLresult} is an intermediate result of Theorem~10 in \cite{KLUCB}.

\vspace{.2in}

\begin{lemma}\label{le:KLpart1}
For any $i\in[N],k\in[M],$ let 
\begin{align*}
    u_{i,k}(t)=\max\{q>z_{i,k}(t):n_{i,k}(t)d(z_{i,k}(t),q)\le(1+\varsigma_i)\delta_i\},
\end{align*} 
where $\delta_i, \varsigma_i$ are arbitrary positive constants. Then, when $n_{i,k}(t)$ is sufficiently large, 
\begin{align*}
    \mathbf{P}(u_{ik}(t)<\mu_k)\le t\exp(-2N\delta_i) + \zeta_i\lceil\delta_i\log t\rceil\exp\Big(-\frac{2N(\delta_i-1)}{3}\Big),
\end{align*}
where $\zeta_i$ is a positive constant.
\end{lemma}

{\bf Proof of Lemma~\ref{le:KLpart1}:}
Since $d(p;q)$ is continuous and increasing for $q\in[p,1],$ the condition $\{u_{ik}(t)<\mu_k\}$ is equivalent to $\{z_{i,k}(t)<\mu_k,\; d(z_{i,k}(t);\mu_k)>(1+\varsigma_i)\delta_i\}$. 
From Lemma~\ref{le:KLresult}, 
\begin{align}
    &\quad\;\mathbf P\bigg(\tilde t_{m-1}<n_{i,k}(t)\le t_m, z_{i,k}(t)<\mu_k, d(z_{i,k}(t);\mu_k)\ge\frac{(1+\varsigma_i)\delta_i}{n_{i,k}(t)}\bigg)\nonumber\\
    &\le \mathbf P\bigg(\lambda(l_m)z_{i,k}(t)-\phi_{\mu_k}(\lambda(l_m))\ge \frac{(1+\varsigma_i)\delta_i}{n_{i,k}(t)(1+\eta_i)}\;,n_{i,k}(t)\in(\tilde t_{m-1}, t_m]\bigg)\nonumber\\
    &= \mathbf P\bigg(\lambda(l_m)\sum_{j\in[N]}\sum_{n=1}^{n_{j,k}(t)}c_{i,j,k,\tau_{j,k,n}}(t)X_{j,k}(\tau_{j,k,n})-\phi_{\mu_k}(\lambda(l_m))\ge \frac{(1+\varsigma_i)\delta_i}{n_{i,k}(t)(1+\eta_i)}, n_{i,k}(t)\in(\tilde t_{m-1}, t_m]\bigg)\nonumber\\
    &\le \mathbf P\bigg(\lambda(l_m)\sum_{j\in[N]}\sum_{n=1}^{n_{j,k}(t)}\frac{1}{Nn_{j,k}(t)}X_{j,k}(\tau_{j,k,n})-\phi_{\mu_k}(\lambda(l_m))\ge \frac{\delta_i}{n_{i,k}(t)(1+\eta_i)}\bigg)\nonumber\\
    &\quad\;+\mathbf P\bigg(\lambda(l_m)\sum_{j\in[N]}\sum_{n=1}^{n_{j,k}(t)}\Big(c_{i,j,k,\tau_{j,k,n}}(t)-\frac{1}{Nn_{j,k}(t)}\Big)X_{j,k}(\tau_{j,k,n})\ge \frac{\varsigma_i\delta_i}{n_{i,k}(t)(1+\eta_i)}, n_{i,k}(t)\in(\tilde t_{m-1}, t_m]\bigg).\label{eq:everyslice}
\end{align}
Let $m^*(\kappa_i)$ be the smallest $m$ such that $t_{m-1} \ge 2F_2(\kappa_i),$ where $\kappa_i$ is a positive constant. 
Unless specified, we consider $m\ge m^*(\kappa_i)$ in the following context. 
Since $d(p;q)$ is continuous and decreasing for $p\in[0,p],$ we have $l_m\ge l_{m^*(\kappa_i)}.$ It is easy to see for $x\le \mu_k,$ it holds that $\lambda(x)\le0$ and that $\lambda(x)$ is increasing. Thus,  $|\lambda(l_m)|\le |\lambda(l_{m^*(\kappa_i)})|$. For the second term of \eqref{eq:everyslice}, for $n_{i,k}(t)\in(\tilde t_{m-1}, t_m],$ then there holds
\begin{align}
    &\quad\;\mathbf P\bigg(\lambda(l_m)\sum_{j\in[N]}\sum_{n=1}^{n_{j,k}(t)}\Big(c_{i,j,k,\tau_{j,k,n}}(t)-\frac{1}{Nn_{j,k}(t)}\Big)X_{j,k}(\tau_{j,k,n})\ge \frac{\varsigma_i\delta_i}{n_{i,k}(t)(1+\eta_i)}\bigg)\nonumber\\
    &\le \mathbf P\bigg(\sum_{j\in[N]}\sum_{n=1}^{n_{j,k}(t)}\Big(c_{i,j,k,\tau_{j,k,n}}(t)-\frac{1}{Nn_{j,k}(t)}\Big)X_{j,k}(\tau_{j,k,n})\le \frac{\varsigma_i\delta_i}{\lambda(l_m)n_{i,k}(t)(1+\eta_i)}\bigg)\nonumber\\
    &\le \mathbf P\bigg(\sum_{j\in[N]}\sum_{h_j=1}^{n_{j,k}(t)}\Big(c_{i,k,j}^{(\tau_{j,h_j})}(t)-\frac{1}{Nn_{j,k}(t)}\Big)X_{j,k}(\tau_{j,h_j})\le \frac{\varsigma_i\delta_i}{\lambda(l_{m^*(\kappa_i)})n_{i,k}(t)(1+\eta_i)}\bigg).\label{cieq}
\end{align}
Let $Y_{j,k}(1), Y_{j,k}(2),\ldots$ be the sub-sequence of $X_{j,k}(\tau)$ for $\tau$ satisfies $a_j(\tau) = k,$ and similarly,  $\{b_{i,j,k}(\cdot)\}$ be the sub-sequence of $c_{i,j,k,n}(t)$ for $\tau$ satisfies $a_j(\tau) = k.$ For any positive integer $N_{jk}$ such that $n_{j,k}(t) = N_{jk},$ it holds that $Y_{j,k}(1) - \mu_k,\ldots, Y_{j,k}(N_{jk})-\mu_k$ are with $0$ mean and bounded support $[-\mu_k, 1-\mu_k]$. Then, from Lemma~\ref{le:Hoff}, for any negative constant $\xi,$ 
\begin{align}\label{bjeq1}
    &\quad\;\mathbf{P}\bigg(\sum_{j\in[N]}\sum_{n_j=1}^{N_{jk}}\Big(b_{i,j,k}(n_j)-\frac{1}{NN_{j,k}}\Big)(Y_{j,k}(n_j)-\mu_k)\le \xi\bigg)\le\exp\bigg(\frac{-2\xi^2}{\sum_{j\in[N]}\sum_{n_j=1}^{N_{jk}}(b_{i,j,k}(n_j)-\frac{1}{NN_{jk}})^2}\bigg).
\end{align}
Since $W$ is doubly stochastic, for any positive integer $\tau,$ there holds $\sum_{j\in[N]}[W^\tau]_{ij}= 1$. Then, from the definition of $b_{i,j,k}(n_j)$,
\begin{align}
    \sum_{j\in[N]}\sum_{n_j=1}^{N_{jk}}b_{i,j,k}(n_j) &= \sum_{j\in[N]}\bigg\{
    \sum_{h=1}^{n_{j,k}(t)-1}\left[W^{t-\tau_{j,h}}-W^{t-\tau_{j,h+1}}\right]_{ij}+\left[W^{t-\tau_{j,n_{j,k}(t)}}\right]_{ij}\bigg\}\nonumber\\
    &= \sum_{h=1}^{n_{j,k}(t)-1}\sum_{j\in[N]}\left[W^{t-\tau_{j,h}}-W^{t-\tau_{j,h+1}}\right]_{ij} + \sum_{j\in[N]}\left[W^{t-\tau_{j,n_{j,k}(t)}}\right]_{ij}
    = 1.\label{bjeq2}
\end{align}
From Lemma~\ref{relation2}, when $n_{i,k}(t)\ge 2F_2(\kappa_i),$ for all $j\in[N],$ there holds \[n_{j,k}(t)\ge \frac{1}{2}n_{i,k}(t)\ge F_2(\kappa_i)\ge  2(M^2+2MN+N).\]
Then, from Lemmas~\ref{relation2} and \eqref{eq:coefficientbound},  
\begin{align}\label{bjeq3}
    \sum_{j\in[N]}\sum_{n_j=1}^{N_{jk}}|b_{i,j,k}(n_j)-\frac{1}{NN_{jk}}|^2\le \sum_{j\in[N]}\sum_{n_j=1}^{N_{jk}}\frac{\kappa_i^2}{N^2}N_{jk}^{-3}\le \frac{\kappa_i^2}{N}N_{jk}^{-2}\le \frac{9\kappa_i^2}{4NN_{ik}^2}.
\end{align}

Let $\xi = \frac{\varsigma_i\delta_i}{\lambda(l_{m^*(\kappa_i)})N_{ik}(t)(1+\eta_i)},$ and $\kappa_i$ be a constant such that 
\begin{align}
    \varsigma_i = \Big|\frac{3\lambda(l_{m^*(\kappa_i)})(1+\eta_i)\kappa_i}{2}\Big|\label{eq:tildevarsigma_i}.
\end{align}
From the definition of $l_m$ in Lemma~\ref{le:KLresult},
it holds that $l_{m^*(\kappa_i)}<\mu_k$. Then, $\lambda(l_{m^*(\kappa_i)})\neq 0$, which together with the fact that $\varsigma_i,\eta_i>0$, implies that $\kappa_i$ uniquely exists.
Then, substituting \eqref{bjeq2} and \eqref{bjeq3} to \eqref{bjeq1}, it holds that
\begin{align*}
    \mathbf{P}\bigg(\sum_{j\in[N]}\sum_{n_j=1}^{N_{jk}}\Big(b_{j,k}(n_j)-\frac{1}{NN_{jk}}\Big)Y_{j,k}(n_j)\le \xi\bigg)\le \exp\bigg(\frac{-2\xi^2}{9\kappa_i^2/(4NN_{ik}^2)}\bigg)\le \exp(-2N\delta_i).
\end{align*}
Then, \eqref{cieq} can be further written as
\begin{align}
    &\quad\;\mathbf P\bigg(\lambda(l_m)\sum_{j\in[N]}\sum_{h_j=1}^{n_{j,k}(t)}\Big(c_{i,k,j}^{(\tau_{j,h_j})}(t)-\frac{1}{Nn_{j,k}(t)}\Big)X_{j,k}(\tau_{j,h_j})\ge \frac{\varsigma_i\delta_i}{n_{i,k}(t)(1+\eta)}, n_{i,k}(t)\in(\tilde t_{m-1}, t_m]\bigg)\nonumber\\
    &\le \sum_{N_{ik} = \tilde t_{m-1}}^{t_m}\mathbf{P}\bigg(\sum_{j\in[N]}\sum_{n_j=1}^{N_{jk}}\Big(b_{j,k}(n_j)-\frac{1}{NN_{jk}}\Big)(Y_{j,k}(n_j)-\mu_k)\le \xi\bigg)\nonumber\\
    &\le (t_m-\tilde t_{m-1})\exp(-2N\delta_i).\label{eq:t^-3}
\end{align}

Next we consider the first term of \eqref{eq:everyslice}. Multiplying $\sum_{l\in[N]}n_{l,k}(t)$ on both sides of the inequality, from the above analysis, $n_{j,k}(t)\ge F_2(\kappa_i)$ for all $j\in[N]$. Then, from Lemma~\ref{relation2}, 
\begin{align}
    &\quad\;\mathbf P\bigg(\lambda(l_m)\sum_{j\in[N]}\sum_{h_j=1}^{n_{j,k}(t)}\frac{1}{Nn_{j,k}(t)}X_{j,k}(\tau_{j,h_j})-\phi_{\mu_k}(\lambda(l_m))\ge \frac{\delta_i}{n_{i,k}(t)(1+\eta_i)}\bigg)\nonumber\\
    &\le \mathbf P\bigg(\lambda(l_m)\sum_{j\in[N]}\sum_{h_j=1}^{n_{j,k}(t)}\frac{\sum_{l\in[N]}n_{l,k}(t)}{Nn_{j,k}(t)}X_{j,k}(\tau_{j,h_j})-\sum_{l\in[N]}n_{l,k}(t)\phi_{\mu_k}(\lambda(l_m))\ge \frac{2N\delta_i}{3(1+\eta_i)}\bigg)\nonumber\\
    &= \mathbf P\bigg(\lambda(l_m)\sum_j\sum_{h_j=1}^{n_{j,k}(t)}\Big(1-\frac{\sum_{l\in[N]}n_{j,k}(t)-n_{l,k}(t)}{Nn_{j,k}(t)}\Big)X_{j,k}(\tau_{j,h_j})-\sum_j n_{j,k}(t)\phi_{\mu_k}(\lambda(l_m))\ge \frac{2N\delta_i}{3(1+\eta_i)}\bigg).\nonumber
\end{align}
Using the fact that all the rewards are within 0 and 1, 
\begin{align*}
    \sum_{j\in[N]}\sum_{h_j=1}^{n_{j,k}(t)}\frac{\sum_{l\in[N]}(n_{j,k}(t)-n_{l,k}(t))}{Nn_{j,k}(t)}X_{j,k}(\tau_{j,h_j})&=\sum_{j\in[N]}\frac{\sum_{l\in[N]}n_{j,k}(t)-n_{l,k}(t)}{N}X_{j,k}(\tau_{j,h_j})\\
    &\le\sum_{j,l:\;n_{j,k(t)>n_{l,k}(t)}}\frac{n_{j,k}(t)-n_{l,k}(t)}{N}
    \le \frac{N(M^2+2MN+N)}{4}.
\end{align*}
Then, 
\begin{align*}
    &\quad\;\mathbf P\bigg(\lambda(l_m)\sum_{j\in[N]}\sum_{h_j=1}^{n_{j,k}(t)}\frac{1}{Nn_{j,k}(t)}X_{j,k}(\tau_{j,h_j})-\phi_{\mu_k}(\lambda(l_m))\ge \frac{\delta_i}{n_{i,k}(t)(1+\eta_i)}\bigg)\nonumber\\
    &\le\mathbf P\bigg(\lambda(l_m)\sum_{j}\sum_{h_j=1}^{n_{j,k}(t)}X_{j,k}(\tau_{j,h_j})-\sum_jn_{j,k}(t)\phi_{\mu_k}(\lambda(l_m))\ge \frac{2N(\delta_i-1)}{3}+\frac{\lambda(l_m)N(M^2+2MN+N)}{4}\bigg).\nonumber
\end{align*}
Note that $X_{j,k}(\tau_{j,h_j})$ are i.i.d. for all $j\in[N]$ and $h_j\le t$. Let $\{\tilde X_k(\tau),\;\tau\in[Nt]\}$ be the combined sequence of $X_{1,k}(0),\ldots,X_{N,k}(0),X_{1,k}(1),\ldots, X_{1,k}(1),\ldots, X_{1,k}(t),\ldots, X_{N,k}(t),$ and
$\{\epsilon_k(\tau)\}$ be the combined sequence $\mathds 1(a_1(0)=k),\ldots, \mathds 1(a_N(0)=k),\ldots,\mathds 1(a_1(t)=k),\ldots,\mathds 1(a_N(t)=k).$
Define \[S_k(t) = \sum_{\tau=0}^t\epsilon_k(\tau)\tilde X_k(\tau),\;\;\; \tilde n_k(t) = \sum_{\tau=0}^t\epsilon_k(\tau).\] 
From Lemma~\ref{le:KLresult}, it holds that
\begin{align*}
    &\quad\;\mathbf P\bigg(\lambda(l_m)\sum_{j\in[N]}\sum_{h_j=1}^{n_{j,k}(t)}\frac{1}{Nn_{j,k}(t)}X_{j,k}(\tau_{j,h_j})-\phi_{\mu_k}(\lambda(l_m))\ge \frac{\delta_i}{n_{i,k}(t)(1+\eta_i)}\bigg)\nonumber\\
    &\le \mathbf{P}\bigg(\lambda(l_m)S_k(t)-\phi_{\mu_k}(\lambda(l_m))\tilde n_k(t)\ge \frac{2N(\delta_i-1)}{3}+\frac{\lambda(l_m)N(M^2+2MN+N)}{4}\bigg)\\
    &\le\exp\Big(-\frac{2N(\delta_i-1)}{3}\Big)\exp\Big(-\lambda(l_m)N(M^2+2MN+N)/4\Big)\\
    &= \exp\Big(-\frac{2N(\delta_i-1)}{3}\Big)\Big(\frac{\mu_k(1-l_m)}{l_m(1-\mu_k)}\Big)^{N(M^2+2MN+N)/4}.
\end{align*}
Substituting the above inequality with \eqref{eq:t^-3} to \eqref{eq:everyslice}, along with the fact that condition $\{u_{ik}(t)<\mu_k\}$ is equivalent to $\{z_{i,k}(t)<\mu_k,\; d(z_{i,k}(t);\mu_k)>(1+\varsigma_i)\delta_i\},$ it follows that
\begin{align*}
   &\quad\; \mathbf P\big(\tilde t_{m-1}<n_{i,k}(t)\le t_m,\; u_{ik}(t)<\mu_k\big)\\
   &\le(t_m-\tilde t_{m-1})\exp(-2N\delta_i) + \exp\Big(-\frac{2N(\delta_i-1)}{3}\Big)\Big(\frac{\mu_k(1-l_m)}{l_m(1-\mu_k)}\Big)^{N(M^2+2MN+N)/4}.
\end{align*}
Let $D_i$ be the first integer such that $t_{D_i}\ge t,$ namely, $D_i=\big\lceil\frac{\log t}{\log(1+\eta_i)}\big\rceil.$ Summing all the $m$ in range up, since 
\[\frac{\mu_k(1-l_{D_i})}{l_{D_i}(1-\mu_k)}\le\frac{\mu_k(1-l_m)}{l_m(1-\mu_k)}\le\frac{\mu_k(1-l_{m^*(\kappa_i)})}{l_{m^*(\kappa_i)}(1-\mu_k)},\]
there exists a $\zeta_i>0$ such that
\begin{align*}
    \sum_{m=m^*(\kappa_i)}^{D_i}\exp\Big(-\frac{2N(\delta_i-1)}{3}\Big)\Big(\frac{\mu_k(1-l_m)}{l_m(1-\mu_k)}\Big)^\frac{N(M^2+2MN+N)}{4}
    &\le \zeta_i\sum_{m=m^*(\kappa_i)}^{D_i}\exp\Big(-\frac{2N(\delta_i-1)}{3}\Big)\\
    &=\zeta_iD_i\exp\Big(-\frac{2N(\delta_i-1)}{3}\Big).
\end{align*}
Since $\log(1+1/(\delta_i - 1))\ge 1/\delta_i,$ the following inequality holds when $n_{i,k}(t)\ge 2F_2(\kappa_i)$:
\begin{align*}
    \mathbf P(u_{ik}(t)<\mu_k)&\le
\sum_{m=m_k^*}^{D_i}\mathbf P\big(\tilde t_{m-1}<n_{i,k}(t)\le t_m,\; u_{ik}(t)<\mu_k\big)
\le t\exp(-2N\delta_i) + \zeta_iD_i\exp\Big(-\frac{2N(\delta_i-1)}{3}\Big)\\
&\le t\exp(-2N\delta_i) + \zeta_i\lceil\delta_i\log t\rceil\exp\Big(-\frac{2N(\delta_i-1)}{3}\Big),
\end{align*}
which completes the proof.
\hfill$\qed$

\vspace{.2in}


\begin{lemma}\label{le:KLpart2}
Define $d^+(x;y)=d(x;y)\mathds 1(x<y)$. Then, it holds that
\begin{align*}
    &\quad\;\;\;\;\sum_{t=0}^T\mathds 1\big(a_i(t)=k,\;n_{i,k}(t)>K_{i,k}(T),\;\mu_1\le u_{i,1}(t),
    \; A_i(t-1)=\emptyset \big)\\
    &\le\sum_{s=K_{i,k}(T)}^T\mathds 1\big(s d^+(z_{i,k}(t);\mu_1)<(1+\varsigma_i)\delta_i,\;n_{i,k}(t)=s\big).
\end{align*}
\end{lemma}

{\bf Proof of Lemma \ref{le:KLpart2}:}
When $\scr A_i(t-1)=\emptyset,$ agent $i$ refers to case (a) of the decision making step of Algorithm~\ref{algorithm:KL}. Note that $a_i(t)=k$ and $\mu_1\le u_{i,1}(t)$ imply that $u_{i,k}(t)\ge u_{i,1}(t)\ge \mu_1,$ 
and hence
\begin{align*}
    d^+(z_{i,k}(t);\mu_1)\le d^+(z_{i,k}(t); u_{i,k}(t))=\frac{(1+\varsigma_i)\delta_i}{n_{i,k}(t)}.
\end{align*}
Then,
\begin{align*}
     &\quad\;\sum_{t=0}^T\mathds 1\big(a_i(t)=k,\;n_{i,k}(t)>K_{i,k}(T),\;\mu_1\le u_{i,1}(t),\; \scr A_i(t-1)=\emptyset \big)\\
     &\le \sum_{t=0}^T\mathds 1\big(a_i(\tau)=k,\;n_{i,k}(t)>K_{i,k}(T),\; n_{i,k}(t)d^+(z_{i,k}(t);\mu_1)\le (1+\varsigma_i)\delta_i\big)\\
     &\le\sum_{t=0}^T\sum_{s=K_{i,k}(T)}^T\mathds 1\big(a_i(t)=k,\;n_{i,k}(t) = s,\;sd^+(z_{i,k}(t);\mu_1)\le (1+\varsigma_i)\delta_i\big)\\
     &= \sum_{s=K_{i,k}(T)}^T\mathds 1\big(n_{i,k}(t) = s,\;sd^+(z_{i,k}(t);\mu_1)\le (1+\varsigma_i)\delta_i\big)\cdot\sum_{t=s}^T\mathds 1\big(a_i(t)=k,\;n_{i,k}(t) = s\big)\\
     &=\sum_{s=K_{i,k}(T)}^T\mathds 1\big(n_{i,k}(t) = s,\;sd^+(z_{i,k}(t);\mu_1)\le (1+\varsigma_i)\delta_i\big),
\end{align*}
which completes the proof.
\hfill$\qed$



Now we are in a position to prove the theorems.

{\bf Proof of Theorem~\ref{thm:KL}:}
00For any $i\in[N], k\in[M],$ it holds that
\begin{align}
    \mathbf{E}(n_{i,k}(T))
    &=\mathbf{E}\bigg(\sum_{t=0}^T\mathds 1(a_i(t)=k)\bigg)\nonumber\\
    &\le K_{i,k}(T)+\mathbf{E}\bigg(\sum_{t=0}^T\mathds 1(a_i(t)=k, n_{i,k}(t)>K_{i,k}(T))\bigg)\nonumber\\
    &\le K_{i,k}(T)+\mathbf{E}\bigg(\sum_{t=0}^T\mathds 1(\mu_{1}>u_{i,1}(t))\bigg)\nonumber\\
    &\quad\,+\mathbf{E}\bigg(\sum_{t=0}^T\mathds 1\big(a_i(t)=k,\;\mu_{1}\le u_{i,1}(t),\; n_{i,k}(t)>K_{i,k}(T),\; A_i(t-1)=\emptyset\big)\bigg)\nonumber\\
&\quad\;+\mathbf{E}\bigg(\sum_{t=0}^T\mathds 1\big(a_i(t)=k,\;n_{i,k}(t)>K_{i,k}(T),\; A_i(t-1)\neq\emptyset\big)\bigg).\label{eq:KL3parts}
\end{align}
We divide the following content into three parts to analyze on the three expectations of the above inequality respectively.

\textbf{Part I:} For the first expectation of \eqref{eq:KL3parts}, there holds 
\begin{align*}
&\quad\;\mathbf{E}\bigg(\sum_{t=0}^T\mathds 1(\mu_{1}>u_{i,1}(t))\bigg)\\
    &\le \mathbf{E}\bigg(\sum_{t=0}^T\mathds 1\big(\mu_{1}>u_{i,1}(t),\;n_{i,1}(t)\ge2F_2(\kappa_i)\big)\bigg)
    +\sum_{t=0}^T\mathbf P\big(\mu_{1}>u_{i,1}(t),n_{i,1}(t)< 2F_2(\kappa_i)\big).
\end{align*}
From the definition of $u_{i,1}(t)$, it holds that  
\begin{align*}
   \sum_{t=0}^T\mathbf P\big(\mu_{1}>u_{i,1}(t),n_{i,1}(t)< 2F_2(\kappa_i)\big)<\sum_{t=0}^T\mathbf P(2F_2(\kappa_i)d(0,\mu_1)\ge Q_i(t)).
\end{align*}
Since $Q_i(t) = O(\log t),$ there exists a constant $\alpha_i,$ such that when $t> \alpha_i-1,$ there holds $2F_2(\kappa_i)d(0,\mu_1)< Q_i(t).$ This suggests that 
\[\sum_{t=0}^T\mathbf P\big(\mu_{1}>u_{i,1}(t),n_{i,1}(t)< 2F_2(\kappa_i)\big)<\alpha_i.\]
From Lemma~\ref{le:KLpart1}, it holds that
\begin{align*}
    &\quad\;\mathbf{E}\bigg(\sum_{t=0}^T\mathds 1\big(\mu_{1}>u_{i,1}(t),\;n_{i,k}(t)\ge2F_2(\kappa_i)\big)\bigg)\\
    &=\sum_{t=0}^T\mathbf{P}\big(\mu_{1}>u_{i,1}(t),\;n_{i,1}(t)\ge2F_2(\kappa_i)\big)\\
    &\le \sum_{t=t(\kappa_i)}^T\bigg(t\cdot e^{-2N\delta_i}+ \zeta_i\lceil\delta_i\log t\rceil\exp\Big(-\frac{2N(\delta_i-1)}{3}\Big)\bigg),
\end{align*}
where $t(\kappa_i)$ stands for the lowest $t$ such that $n_{i,1}(t)\ge 2F_2(\kappa_i).$ From the definition of $F_2(\cdot)$ in Remark~\ref{re:def}, it holds that $t(\kappa_i)\ge F_2(\kappa_i)\ge 2(M^2+2MN+N).$ 
Then, substituting $\delta_i=3(\log t+3\log\log t)/(2|\scr N_i|),$ which is defined in Theorem~\ref{thm:KL}, to the above inequality, we obtain
\begin{align*}
    &\quad\;\mathbf{E}\bigg(\sum_{t=0}^T\mathds 1\big(\mu_{1}>u_{i,1}(t),\;n_{i,k}(t)\ge2F_2(\kappa_i)\big)\bigg)\\
    &\le \sum_{t=t(\kappa_i)}^Tt^{-2}+\sum_{t=t(\kappa_i)}^T\frac{3\zeta_ie^{\frac{2N}{3}}}{2|\scr N_i|}\frac{\log^2t+3\log t\log\log t}{(t\log^3 t)^{\frac{N}{|\scr N_i|}}}\\
    &\le \frac{\pi^2}{6}+\sum_{t=t(\kappa_i)}^T\frac{3\zeta_ie^{\frac{2N}{3}}}{2|\scr N_i|}\frac{\log^2t+3\log t\log\log t}{(t\log^3 t)^{\frac{N}{|\scr N_i|}}}.
\end{align*}
Note that
$$\frac{\log^2t+3\log t\log\log t}{(t\log^3 t)^{N/|\scr N_i|}}=O\Big(\frac{1}{t\log t}\Big).$$ There exists a $\Xi_i>0,$ such that the last term of the above inequality is bounded by $\Xi_i\log\log T,$ where $\Xi_i$ is upper bounded by $3\zeta_ie^{\frac{2N}{3}}/(2|\scr N_i|).$
Thus,
\begin{align}
    \mathbf{E}\bigg(\sum_{t=0}^T\mathds 1(\mu_{1}>u_{i,1}(t))\bigg)\le \alpha_i+\frac{\pi^2}{6}+\Xi_i\log\log T.\label{eq:finalKLpart1}
\end{align}

\textbf{Part II:} For the second expectation of \eqref{eq:KL3parts}, we obtain from Lemma~\ref{le:KLpart2} that
\begin{align*}
    &\quad\;\mathbf{E}\bigg(\sum_{t=0}^T\mathds 1\big(a_i(t)=k,\;\mu_{1}\le u_{i,1}(t),\; n_{i,k}(t)>K_{i,k}(T),
    \;A_i(t-1)=\emptyset\big)\bigg)\\
    &=\sum_{t=0}^T\mathbf{P}\big(a_i(t)=k,\;\mu_{1}\le u_{i,1}(t),\; n_{i,k}(t)>K_{i,k}(T),\; A_i(t-1)=\emptyset\big)\\
    &\le \sum_{s=K_{i,k}(T)}^T\mathbf{P}\big(s d^+(z_{i,k}(t);\mu_1)\le(1+\varsigma_i)\delta_i,\;n_{i,k}(t)=s\big)\\
    &\le \sum_{s=K_{i,k}(T)}^T\mathbf{P}\big( d^+(z_{i,k}(t);\mu_1)\le\frac{(1+\varsigma_i)\delta_i}{K_{i,k}(T)},\;n_{i,k}(t)=s\big)\\
    &\le \sum_{s=K_{i,k}(T)}^T\mathbf{P}\Big( d^+(z_{i,k}(t);\mu_1)\le\frac{d(\mu_k;\mu_1)}{1+\epsilon},\;n_{i,k}(t)=s\Big).
\end{align*}
Let $r_k(\epsilon)\in(\mu_k,\mu_1)$ be such that $d(r_k(\epsilon),\mu_1)=d(\mu_k;\mu_1)/(1+\epsilon)$. Then, if $d^+(z_{i,k}(t);\mu_1)\le\frac{d(\mu_k;\mu_1)}{1+\epsilon}$, we have $z_{i,k}(t)\ge r_k(\epsilon)>\mu_k$, then
\begin{align*}
   \mathbf{P}\Big( d^+(z_{i,k}(t);\mu_1)\le\frac{d(\mu_k;\mu_1)}{1+\epsilon},\;n_{i,k}(t)=s\Big)
    \le\mathbf{P}\big(d(z_{i,k}(t);\mu_k)\ge d(r_k(\epsilon);\mu_k),\;n_{i,k}(t)=s\big).
\end{align*}
Since $n_{i,k}(t)\ge 2F_2(\kappa_i),$ from the proof of Lemma~\ref{le:KLpart1}, it follows that
\begin{align*}
    \mathbf{P}\Big( d^+(z_{i,k}(t);\mu_1)\le\frac{d(\mu_k;\mu_1)}{1+\epsilon},\;n_{i,k}(t)=s\Big)
    \le \zeta_i\exp\Big(-\frac{2Nd(r_k(\epsilon);\mu_k)}{3(1+\varsigma_i)}s\Big).
\end{align*}
Let $C_3(\epsilon,\varsigma_i)=\frac{2Nd(r_k(\epsilon);\mu_k)}{3(1+\varsigma_i)}$. Then,
\begin{align*}
    &\quad\;\sum_{s=K_{i,k}(T)}^T\mathbf{P}\Big( d^+(z_{i,k}(t);\mu_1)\le\frac{d(\mu_k;\mu_1)}{1+\epsilon},\;n_{i,k}(t)=s\Big)\\
    &\le\sum_{s=K_{i,k}(T)}^T\zeta_i\exp(-C_3(\epsilon,\varsigma_i)s)\\
    &\le\frac{\zeta_i\exp(-C_3(\epsilon,\varsigma_i)K_{i,k}(T))}{1-\exp(-C_3(\epsilon,\varsigma_i))}\\
    &\le \frac{\zeta_iC_2(\epsilon,\varsigma_i)}{T^{\gamma(\epsilon,\varsigma_i)}},
\end{align*}
where 
\begin{align*}
    C_2(\epsilon,\varsigma_i) = \frac{1}{1-\exp(-C_3(\epsilon,\varsigma_i))},\hspace{.3in}
    \gamma(\epsilon,\varsigma_i) = \frac{3(1+\varsigma_i)(1+\epsilon)C_3(\epsilon,\varsigma_i)}{2|\scr N_i|d(\mu_k;\mu_1)}.
\end{align*}
Easy computation shows that $r_k(\epsilon)-\mu_k = O(\epsilon)$. Thus, from Pinsker's inequality, $C_3(\epsilon,\varsigma_i)=O(\epsilon^2),$ which implies that $C_2(\epsilon,\varsigma_i)=O(\epsilon^{-2})$ and that $\gamma(\epsilon,\varsigma_i)=O(\epsilon^2).$

\textbf{Part III:}
The last expectation in \eqref{eq:KL3parts} stands for the expected number of pulls of agent $i$ on arm $k$ in case (b) in the decision making step after time $t_1,$ where $t_1\triangleq\argmin_t\{n_{i,k}(t-1)= K_{i,k}(T)\},$ while the sum of first two expectations upper bounds that in case (a). And the last expectation can be understood as the extra pulls an agent makes in order to ``catch up'' with the largest pull among all the agents on arm $k$. In this sense, it is easy to see that any pull made in case (b) would not affect the global maximal pull, while any pull made in case (a) would increase the global maximal pull by at most 1. With this is mind, let $i_1=\argmax_i n_{i,k}(t_1)$ be the agent who makes the most pulls on arm $k$ at $t_1,$ and $g_{i,k}$ be the number of pulls made by $i$ on arm $k$ in case (a) after $t_1$. Then, for all $i\in[N],$ it holds that
\begin{align*}
    n_{i,k}(T)\le \max_j n_{j,k}(T) \le n_{i_1,k}(t_1) + \sum_{j\in[N]}g_{j,k}.
\end{align*}
From Lemma~\ref{relation2}, 
$n_{i_1,k}(t_1) -n_{i,k}(t_1)\le M^2 + 2MN +M,$ and thus
\begin{align*}
    &\mathbf{E}\bigg(\sum_{t=0}^T\mathds 1\big(a_i(t)=k,\;n_{i,k}(t)>K_{i,k}(T),\; \scr A_i(t-1)\neq\emptyset\big)\bigg)\\
    =\;&\mathbf{E}\big(n_{i,k}(T)-n_{i,k}(t_1)-g_{i,k}\big)\\
    \le\;& n_{i_1,k}(t_1) -n_{i,k}(t_1) + \sum_{[N]\setminus\{i\}}\mathbf{E}(g_{j,k})\\
    \le\;& M^2 + 2MN +M+ \sum_{[N]\setminus\{i\}}\mathbf{E}(g_{j,k}).
\end{align*}
From the analysis in part I and part II, for any $j\in[N],$ it holds that
\begin{align*}
    \mathbf{E}(g_{j,k})\le \alpha_j+\frac{\pi^2}{6}+\Xi_j\log\log T+\frac{\zeta_iC_2(\epsilon,\varsigma_j)}{T^{\gamma(\epsilon,\varsigma_j)}}.
\end{align*}
Then, 
\begin{align*}
   &\quad\mathbf{E}\bigg(\sum_{t=0}^T\mathds 1\big(a_i(t)=k,\;n_{i,k}(t)>K_{i,k}(T),\; \scr A_i(t-1)\neq\emptyset\big)\bigg)\\
    &\le \sum_{j\in[N]\setminus\{i\}}\Big(\alpha_j+\frac{\pi^2}{6}+\Xi_j\log\log T+\frac{\zeta_jC_2(\epsilon,\varsigma_j)}{T^{\gamma(\epsilon,\varsigma_j)}}\Big).
\end{align*}
Let 
\begin{align*}
    \Psi(\epsilon, T)
    &=M^2 + 2MN +M+\sum_{j\in[N]}\Big(\alpha_j+\frac{\pi^2}{6}+\Xi_j\log\log T+\frac{\zeta_jC_2(\epsilon,\varsigma_j)}{T^{\gamma(\epsilon,\varsigma_j)}}\Big).
\end{align*}
Combining the results of all three parts, 
$
    \mathbf{E}(n_{i,k}(T))\le K_{i,k}(T) + \Psi(\epsilon, T)$.
Then, 
\begin{align*}
    R_i(T)&=\sum_{k:\Delta_k>0}\mathbf{E}(n_{i,k}(T))\Delta_k
    \le \sum_{k:\Delta_k>0} (K_{i,k}(T) + \Psi(\epsilon, T))\Delta_k,
\end{align*}
which completes the proof.
\hfill$\qed$

{\bf Proof of Theorem~\ref{thm:editedmaintheorem}:}
Let \[L = \max\bigg\{\frac{12(1+\beta_i)^2\log T}{|\scr N_i|\Delta_{k}^2},2F_2(\beta_i)\bigg\}.\]
From the decision making step of the algorithm,
\begin{align*}
    n_{i,k}(T)&=1+\sum_{t=1}^T\mathds 1(a_i(t)=k)\\
    &\le L+\sum_{t=1}^T\mathds 1(a_i(t)=k, n_{i,k}(t-1)\ge L)\\
    &\le \sum_{t=1}^T\mathds 1\Big(z_{i,k}(t)+C(t,n_{i,k}(t))\ge z_{i,1}(t)+C(t,n_{i,1}(t)),n_{i,k}(t-1)\ge L\Big)+L\\
    &\quad\;+\sum_{t=1}^T\mathds 1(a_i(t)=k, k\in\scr A_i(t-1), n_{i,k}(t-1)\ge L).
\end{align*}
Thus,
\begin{align}
        \mathbf{E}(n_{i,k}(T))
        \le\;& L+\sum_{t=1}^T\mathbf{P}\Big(z_{i,k}(t)+C(t,n_{i,k}(t))\ge z_{i,1}(t)  
        +C(t,n_{i,1}(t)), n_{i,k}(t-1)\ge L,a_i(t)=k\Big)\nonumber\\
    &+\mathbf{E}\bigg(\sum_{t=1}^T\mathds 1(a_i(t)=k, k\in\scr A_i(t-1), n_{i,k}(t-1)\ge L)\bigg),\label{eq:totalestimate}
\end{align}
of which, the second and third term stand for the number of pulls made in case (a) and case (b) of the decision making step after agent $i$ pulls $L$ times of arm $k$ respectively. We divide the following analysis into two parts to estimate the two terms~separately.

\textbf{Part A:}
For the second term of \eqref{eq:totalestimate}, 
\begin{align}
    &\quad\,\sum_{t=1}^T\mathbf{P}\Big(z_{i,k}(t)+C(t,n_{i,k}(t))\ge z_{i,1}(t)    +C(t,n_{i,1}(t)),  n_{i,k}(t-1)\ge L,a_i(t)=k\Big)\nonumber\\
        &=\sum_{t=1}^T\sum_{N_{i1}=1}^{t-1}\sum_{N_{ik}=L}^{t-1}\mathbf{P}\Big(z_{i,k}(t)+C(t,n_{i,k}(t))\ge z_{i,1}(t) +C(t,n_{i,1}(t)), n_{i,k}(t-1)=N_{ik},\;n_{i,1}(t-1)=N_{i1}\Big)\nonumber\\
        &\le \bigg(\sum_{t=1}^T\sum_{N_{i1}=2F_2(\beta_i)}^{t-1}\sum_{N_{ik}=L}^{t-1}+\sum_{t=1}^T\sum_{N_{i1}=1}^{2F_2(\beta_i) - 1}\sum_{N_{ik}=L}^{t-1}\bigg) \nonumber\\
        &\qquad \mathbf{P}\Big(z_{i,k}(t)+C(t,n_{i,k}(t))\ge z_{i,1}(t)+C(t,n_{i,1}(t)),\;n_{i,k}(t-1)=N_{ik},n_{i,1}(t-1)=N_{i1}\Big)\nonumber\\
        &\le\sum_{t=1}^T\sum_{N_{i1}=2F_2(\beta_i)}^{t-1}\sum_{N_{ik}=L}^{t-1}\mathbf{P}\Big(z_{i,k}(t)+C(t,n_{i,k}(t))\ge z_{i,1}(t)\nonumber\\
        &\qquad\qquad\qquad\qquad\quad\quad\quad\; +C(t,n_{i,1}(t)), n_{i,k}(t-1)=N_{ik},\;n_{i,1}(t-1)=N_{i1}\Big)\nonumber\\
        &\quad\,+\sum_{t=1}^T\mathbf{P}\big(C(t,n_{i,k}(t))+1>C(t,n_{i,1}(t)),n_{i,1}(t-1)\le 2F_2(\beta_i) - 1,n_{i,k}(t-1)\ge L\big).\label{3sum}
\end{align}
First consider the last summation term of the above inequality, there holds
\begin{align*}
    &\quad\;\sum_{t=1}^T\mathbf{P}\big(C(t,n_{i,k}(t))+1>C(t,n_{i,1}(t)),n_{i,1}(t-1)\le 2F_2(\beta_i) - 1,n_{i,k}(t-1)\ge L\big)\\
    &\le \sum_{t=1}^T\mathds 1\big(C(t,L)+1>C(t,2F_2(\beta_i))\big).
\end{align*}
It is easy to see that there exists a constant $\alpha_i',$ such that when $t>\alpha_i',$ it holds that $C(t,L)+1< C(t,2F_2(\beta_i)).$ Consequently,
\begin{align}\label{arm1restrict}
    \sum_{t=1}^T\mathbf{P}\big(C(t,n_{i,k}(t))+1>C(t,n_{i,1}(t)),n_{i,1}(t-1)\le 2F_2(\beta_i) - 1,n_{i,k}(t-1)\ge L\big)\le \alpha_i'.
\end{align}
For each term in the first summation of \eqref{3sum}, it can be divided into three cases,
\begin{equation}\label{threecases}
    \begin{split}
        &\quad\,\mathbf{P}\Big(z_{i,k}(t)+C(t,n_{i,k}(t)\ge z_{i,1}(t)+C(t,n_{i,1}(t)),\; n_{i,k}(t-1)=N_{ik},n_{i,1}(t - 1)=N_{i1}\Big)\\
    &\le \mathbf{P}(z_{i,k}(t)-\mu_{k}\geq C(t,n_{i,k}(t)),n_{i,k}(t-1)=N_{ik})\\
    &\quad\;+\mathbf{P}(\mu_{1}-z_{i,1}(t)\geq C(t,n_{i,1}(t)),n_{i,1}(t - 1)=N_{i1})\\
    &\quad\;+\mathbf{P}(\mu_{1}-\mu_{k}< 2C(t,n_{i,k}(t)),n_{i,k}(t-1)=N_{ik}),
    \end{split}
\end{equation}
where $N_{ik} \ge L \ge 2F_2(\beta_i)$ and $N_{i1} \ge F_2(\beta_i).$
The technique of solving the first two inequalities are exactly the same, so we will only work on the former one in the following context. Note that $z_{i,k}(t) = \sum_{j\in[N]}\sum_{\tau=1}^t c_{i,k,j}^{(\tau)}(t)X_{j,k}(\tau),$ and that $\sum_{j\in[N]}\sum_{\tau=1}^t c_{i,k,j}^{(\tau)}(t)=1$, we can re-express $z_{i,k}(t)-\mu_{k}$ as
\begin{equation}\label{z-mu}
    \begin{split}
        z_{i,k}(t)-\mu_{k} &= \sum_{j\in[N]}\sum_{\tau=1}^t c_{i,k,j}^{(\tau)}(t)(X_{j,k}(\tau)-\mu_{k})
    \end{split}
\end{equation}
For $\{Y_{j,k}(\tau)\}, \{b_{i,j,k}(\cdot)\}$ defined in the proof of Lemma~\ref{le:KLpart1}, and any constant $N_{ik}$ such that $n_{i,k}(t)=N_{ik},$ the sequence $Y_{i,k}(1)-\mu_{k},\ldots,Y_{i,k}(N_{ik})-\mu_{k}$ is with 0 mean and bounded range $[-\mu_{k},1-\mu_{k}]$. Thus, from \eqref{Hoeffding2}, for any positive $\eta$, 
\begin{equation}\label{APPLYHOEFF}
    \begin{split}
        \mathbf{P}\bigg(\sum_{j\in[N]}\sum_{n_j=1}^{N_{jk}}b_{i,j,k}(n_j)(Y_{j,k}-\mu_{k})\ge \eta\bigg)
        \le\exp\bigg(-\frac{2\eta^2}{\sum_{j\in[N]}\sum_{n_j=1}^{N_{jk}}b_{i,j,k}(n_j)^2}\bigg).
    \end{split}
\end{equation}
From Lemma~\ref{relation2}, when $n_{i,k}(t-1)=N_{ik},$ we have $n_{j,k}(t-1)\in[\frac{N_{ik}}{2},\frac{3N_{ik}}{2}]$. For any decreasing function $h(x)$, let $\scr B = \big\{N_{jk}\in[\frac{N_{ik}}{2},\frac{3N_{ik}}{2}], \forall j\in[N]\big\},$ and \[\scr C = \bigg\{\sum_{j\in[N]}\sum_{n_j=1}^{N_{jk}} b_{i,j,k}(n_j)( Y_{j,k}(n_j)-\mu_{k})\ge h(N_{ik})\bigg\}.\]  Then, for $N_{ik}\ge L\ge 2F_2(\beta_i),$ 
\begin{align*}
    &\quad\mathbf{P}\bigg(\sum_{j\in[N]}\sum_{\tau=1}^t c_{i,j,k,n}(t)( X_{j,k}(\tau)-\mu_{jk})\ge h(n_{i,k}(t)),\; n_{i,k}(t-1)=N_{ik}\bigg)\\
    &\le\max_{\scr B}\mathbf{P}(\scr C, n_{i,k}(t-1)= N_{ik})\\
    &\le\max_{\scr B}\mathbf{P}(n_{i,k}(t-1)=N_{ik}|\;\scr C)\cdot\mathbf{P}(\scr C)\\
    &\le \max_{\scr B}\mathbf{P}(\scr C).
\end{align*}
 Let $h(N_{ik})=(1+\beta_i)\sqrt{\frac{3\log t}{|\scr N_i|N_{ik}}}$. From \eqref{APPLYHOEFF}, there holds
\begin{align*}
 \max_{\scr B}\mathbf{P}(\scr C)&\le \max_{\scr B}\exp\bigg(-\frac{2h^2(N_{ik})}{\sum_{j\in[N]}\sum_{n_j=1}^{N_{jk}} b^2_{i,j,k}(n_j)}\bigg)
    \le\max_{\scr B} \exp\bigg(-\frac{\frac{6(1+\beta_i)^2\log t}{NN_{ik}}}{\sum_{j\in[N]}\sum_{n=1}^{N_{jk}}\frac{(1+\beta_i)^2}{N^2N_{jk}^2}}\bigg).
\end{align*}
Since  $W_\infty=\frac{1}{N}\1\1'$ when $\bbb G$ is undirected and connected, we obtain from Lemma~\ref{le:distance} that 
\begin{align}\label{eq:sumofW}
    \sum_{j\in[N]}\sum_{n_j=1}^{N_{jk}} b^2_{i,j,k}(n_j)\le \sum_{j\in[N]}\sum_{n=1}^{N_{jk}}\frac{(1+\beta_i)^2}{N^2N_{jk}^2}
    =\frac{(1+\beta_i)^2}{NN_{jk}},
\end{align}
then using Lemma~\ref{relation2}, when $N_{ik}\ge 2F_2(\beta_i),$ it holds that
\begin{align*}
    \max_{\scr B}\mathbf{P}(\scr C)
    \le\max_{\scr B}\exp\bigg(-\frac{6N}{\sum_{j\in[N]}\frac{N_{ik}}{N_{jk}}}\log t\bigg)
    \le\exp\bigg(-\frac{6 N}{3N/2}\log t\bigg)\le t^{-4}.
\end{align*}
Thus, for $N_{ik}\ge 2F_2(\beta_i),$ it holds that
\begin{align}\label{XmuHoeff}
    \mathbf{P}\bigg(\sum_{j\in[N]}\sum_{\tau=1}^t c_{i,k,j}^{\tau}(t)( X_{j,k}(\tau)-\mu_{jk})\ge h(n_{i,k}(t)),\; n_{i,k}(t-1)=N_{ik}\bigg)\le t^{-4}.
\end{align}

Now let \[C(t,n_{i,k}(t))=(1+\beta_i)\sqrt{\frac{3\log t}{|\scr N_i|n_{i,k}(t)}}.\] Combining \eqref{XmuHoeff} and \eqref{z-mu}, for $N_{ik}\ge 2F_2(\beta_i),$ 
\begin{align}
    &\quad\;\mathbf{P}\big(z_{i,k}(t)-\mu_{k}\geq C(t,n_{i,k}(t)),n_{i,k}(t-1)=N_{ik}\big)\nonumber\\
    &\le\mathbf{P}\bigg(\sum_{j\in[N]}\sum_{\tau=1}^t c_{i,k,j}^{(\tau)}(t)( X_{j,k}(\tau)-\mu_{k})\ge h(N_{ik}),\; n_{i,k}(t-1)= N_{ik}\bigg)
    \le t^{-4}\nonumber.
\end{align}
Similarly, for $N_{i1}\ge 2F_2(\beta_i),$ 
\begin{align*}
    \mathbf{P}\big(\mu_{1}-z_{i,1}(t)\geq C(t,n_{i,1}(t)),n_{i,1}(t-1)=N_{i1}\big)\le t^{-4}.
\end{align*}
Then, consider the last term of \eqref{threecases}. It is easy to verify when $n_{i,k}(t)\ge \frac{12(1+\beta_i)^2\log T}{|\scr N_i|\Delta_{k}^2},$ it always holds that $\mu_1-\mu_k>2C(t,n_{i,k}(t)).$
Substituting these results to \eqref{threecases}, we obtain when $N_{ik}, N_{i1}\ge 2F_2(\beta_i),$
\begin{align}\label{eq:undirected_1/t^4}
    &\mathbf{P}\Big(z_{i,k}(t)+C(t,n_{i,k}(t))\ge z_{i,1}(t)+C(t,n_{i,1}(t)), \; n_{i,k}(t-1)=N_{ik},n_{i,1}(t - 1)=N_{i1}\Big)\le 2t^{-4}.
\end{align}
Combining \eqref{eq:undirected_1/t^4} with \eqref{3sum} and \eqref{arm1restrict}, 
\begin{align}
    &\quad\,\sum_{t=1}^T\mathbf{P}\Big(z_{i,k}(t)+C(t,n_{i,k}(t))\ge z_{i,1}(t)+C(t,n_{i,1}(t)),\; n_{i,k}(t-1)\ge L\Big)\nonumber\\
    &\le \sum_{t=1}^T\sum_{N_{i1}=2F_2(\beta_i)}^{t-1}\sum_{N_{ik}=L}^{t-1} 2t^{-4}+\alpha_i'\nonumber\\
        &\le\frac{\pi^2}{3}+\alpha_i'\label{small}.
\end{align}

\textbf{Part B:}
Now what is left is to estimate the last term of \eqref{eq:totalestimate}, which stands for the expected number of pulls of agent $i$ on arm $k$ in case (b) in the decision making step after time $t_1,$ where $t_1\triangleq\argmin_t\{n_{i,k}(t-1)= L\},$ and can be intuitively understood as the extra pulls an agent makes in order to ``catch up'' with the largest pull among all the agents on arm $k$. In this sense, it is easy to see that any pull made in case (b) would not affect the global maximal pull, while any pull made in case (a) would increase the global maximal pull by at most 1. With this in mind, let $i_1=\argmax_i n_{i,k}(t_1)$ be the agent who makes the most pulls on arm $k$ at $t_1,$ and $g_{i,k}$ be the number of pulls made by $i$ on arm $k$ in case (a) after $t_1$. Then, for all $i\in[N],$ it holds that
\begin{align*}
    n_{i,k}(T)\le \max_j n_{j,k}(T) \le n_{i_1,k}(t_1) + \sum_{j\in[N]}g_{j,k}.
\end{align*}
Thus, 
\begin{align*}
    &\quad\;\mathbf{E}\bigg(\sum_{t=1}^T\mathds 1(a_i(t)=k, k\in\scr A_i(t-1), n_{i,k}(t-1)\ge L)\bigg)\\
    &= \mathbf{E}(n_{i,k}(T) - n_{i,k}(t_1)- g_{i,k})\\
    &\le \mathbf{E}(n_{i_1,k}(t_1) - n_{i,k}(t_1)) + \mathbf{E}\bigg(\sum_{j\in[N]}g_{j,k} - g_{i,k}\bigg).
\end{align*}

From \eqref{small}, we have $\mathbf{E}(g_{j,k}) = \frac{\pi^2}{3}+\alpha_j'$ for all $j\in[N].$ From Lemma~\ref{relation2}, 
$\mathbf{E}(n_{i_1,k}(t_1) - n_{i,k}(t_1))\le M^2+2MN+N$. With these together, 
\begin{align*}
     &\quad\;\mathbf{E}\bigg(\sum_{t=1}^T\mathds 1(a_i(t)=k, k\in\scr A_i(t-1), n_{i,k}(t-1)\ge L)\bigg)\\ 
     &\le M^2+2MN+N + \sum_{j\in[N]\setminus\{i\}}\bigg(\frac{\pi^2}{3}+\alpha_j'\bigg).
\end{align*}
Combining the result with \eqref{eq:totalestimate} and \eqref{small}, 
\begin{align*}
    \mathbf{E}(n_{i,k}(T))\le L+ M^2+2MN+N
    + \sum_{j\in[N]}\bigg(\frac{\pi^2}{3}+\alpha_j'\bigg).
\end{align*}
Let \[\Gamma = M^2+2MN+N
    + \sum_{j\in[N]}\bigg(\frac{\pi^2}{3}+\alpha_j'\bigg),\] then
\begin{align*}
    R_i(T)&=\sum_{k:\Delta_k>0}\mathbf{E}(n_{i,k}(T))\Delta_k\le \sum_{k:\Delta_k>0}(L+ \Gamma)\Delta_k,
\end{align*}
which completes the proof.
\hfill$\qed$

{\bf Proof of Theorem~\ref{thm:directed}:} The analysis follows the same procedure as in the proof of Theorem~\ref{thm:editedmaintheorem}, except when dealing with \eqref{eq:sumofW}, due to the difference in weight matrix $W$ used in the update, it holds in this setting that
\begin{align*}
    \sum_{j\in[N]}\sum_{n_j=1}^{N_{jk}} b^2_{i,j,k}(n_j)\le \sum_{j\in[N]}\sum_{n=1}^{N_{jk}}\frac{(1+\beta_i)^2[W_\infty]^2_{ij}}{N_{jk}^2}
    =\sum_{j\in[N]}\frac{(1+\beta_i)^2[W_\infty]^2_{ij}}{N_{jk}}.
\end{align*}
From Lemma~\ref{relation2}, when $N_{ik}\ge 2F_2(\beta_i)$, it holds that $N_{jk}\ge \frac{2}{3}N_{ik}$ for all $i,j\in[N].$ Then the above inequality can be further bounded as 
\begin{align}\label{eq:bbound_direct}
    \sum_{j\in[N]}\sum_{n_j=1}^{N_{jk}} b^2_{i,j,k}(n_j)&\le \sum_{j\in[N]}\frac{3(1+\beta_i)^2[W_\infty]^2_{ij}}{2N_{ik}}.
\end{align}
From Lemma~\ref{le:directrho2}, it holds that
\begin{align}
    \sum_{j=1}^N[W_{\infty}]_{ij}^2=a^{\top} a= a^{\top} W a = {\rm tr}(a^{\top} W a)={\rm tr}(W\cdot aa^{\top})
    \le \sum_{i=1}^N \frac{\sum_{j=1}^N a_ia_j}{|\scr N_i|}\nonumber
\end{align}
where we make use of the property that ${\rm tr}(AB)={\rm tr}(BA)$ and $a$ is positive. Here,
${\rm tr}(\cdot)$ denotes the trace of a square matrix. From Lemma~\ref{le:directrho2}, $\1'a=1$, then
\begin{align}\label{sumofsquares}
    \sum_{j=1}^N[W_{\infty}]_{ij}^2&\le \sum_{i=1}^N\frac{a_i}{|\scr N_i|}\le\sum_{i=1}^N\frac{a_i}{2}=\frac{1}{2}.
\end{align}
Substituting the above inequality to \eqref{eq:bbound_direct}, we obtain that 
\begin{align*}
    \sum_{j\in[N]}\sum_{n_j=1}^{N_{jk}} b^2_{i,j,k}(n_j)&\le \frac{3(1+\beta_i)^2}{4N_{ik}}.
\end{align*}
Let $h(N_{ik}) = (1+\beta_i)\sqrt{\frac{3\log t}{2N_{ik}}}$. Using Lemma~\ref{le:Hoff}, we obtain that
\begin{align}
    \max_{\scr B}\mathbf{P}(\scr C)
    \le\max_{\scr B}\exp\bigg(-\frac{2h^2(N_{ik})}{\sum_{j\in[N]}\sum_{n_j=1}^{N_{jk}} b^2_{i,j,k}(n_j)}\bigg)
    \le t^{-4}, \nonumber
\end{align}
where $\scr B$ and $\scr C$ are defined in the proof of Theorem~\ref{thm:editedmaintheorem}. Similar to the analysis in the proof of Theorem~\ref{thm:editedmaintheorem}, we obtain that 
when $N_{ik}, N_{i1}\ge 2F_2(\beta_i),$
\begin{align}\label{eq:directed_1/t^4}
    \mathbf{P}\Big(z_{i,k}(t)+C(t,n_{i,k}(t))\ge z_{i,1}(t)+C(t,n_{i,1}(t)), \; n_{i,k}(t-1)=N_{ik},n_{i,1}(t - 1)=N_{i1}\Big)\le 2t^{-4}.
\end{align}
Then following the remaining procedure as in the proof of Theorem~\ref{thm:editedmaintheorem}, we complete the proof.
\hfill$\qed$

\section{Simulations} \label{sec:simulations}
Numerical experiments carried out with the help of the libraries in~\cite{harris2020array, 2020SciPy-NMeth, Hunter:2007, SciPyProceedings_11, mckerns-proc-scipy-2011} are presented here 
to empirically evaluate and compare the performance of our proposed algorithms. We first contrast the performance of each algorithm with its single agent counterpart, with different choice values for the parameters $\varsigma$ and $\beta$, dropping the subscript temporarily since we hold them constant across agents for now. Specifically for decentralized UCB1, we conduct performance comparisons over (1) undirected and connected graphs (2) directed and strongly connected graphs.
We then pit the two decentralized algorithms against each other for $\varsigma=\beta=0.01$. Finally, we compare the performance of our decentralized UCB1 algorithm (with $\beta$ being chosen as 1, 0.1, or 0.01)  and the algorithm in \cite{jingxuan2021}.

\begin{figure}[!tb]
    \centering
    \includegraphics[width=0.8\textwidth]{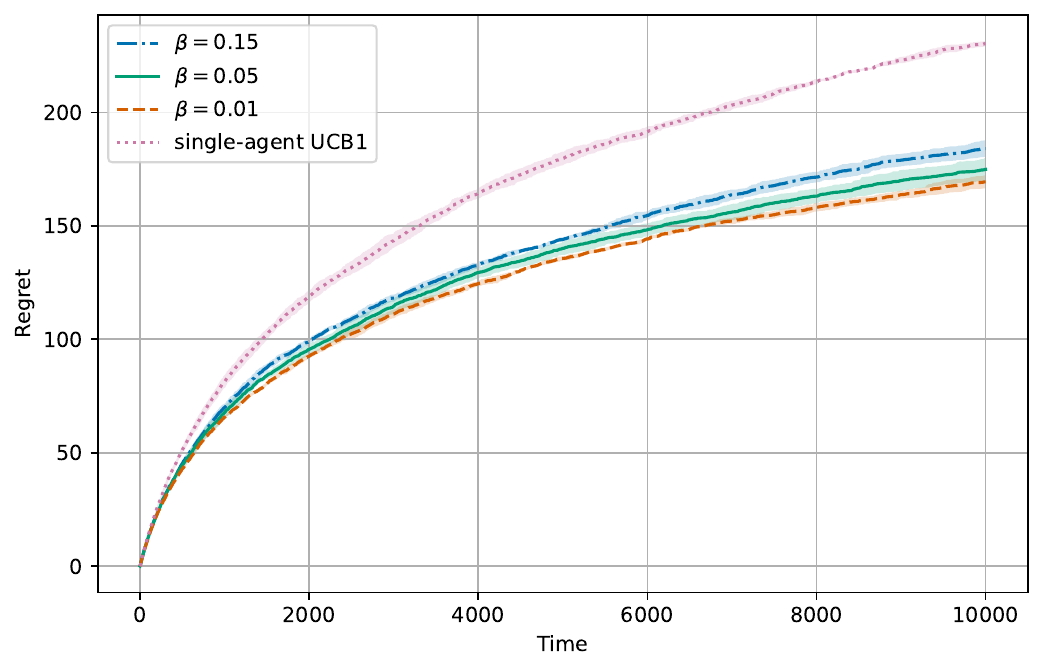}
    \caption{UCB1 vs. decentralized UCB1 over directed graphs}
    \label{fig:directed_normal}
\end{figure}

\begin{figure}[!tb]
    \centering
    \includegraphics[width=0.8\textwidth]{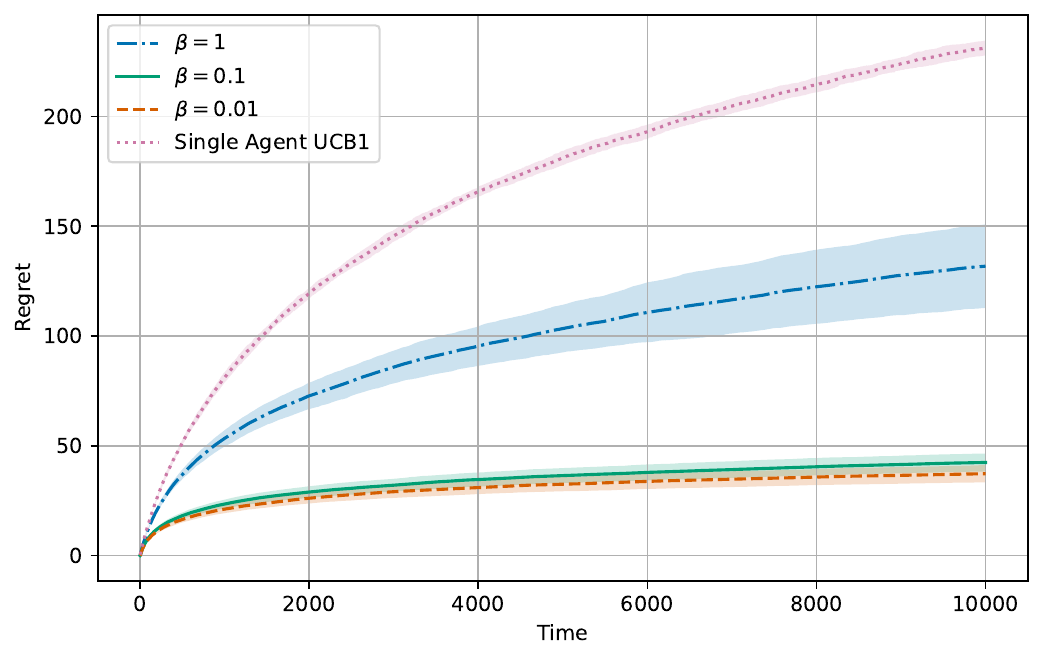}
    \caption{UCB1 vs. decentralized UCB1}
    \label{fig:dist_UCB1_vs_single_UCB1}
\end{figure}

\begin{figure}[!tb]
    \centering
    \includegraphics[width=0.8\textwidth]{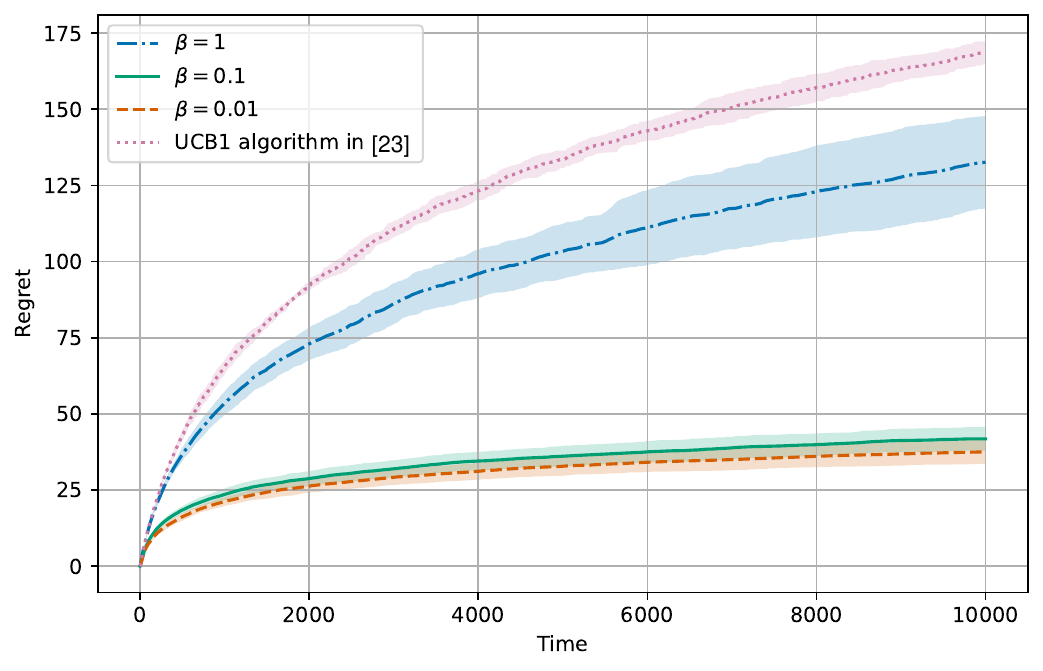}
    \caption{Plots of the mean regrets of decentralized UCB1 in our paper vs. UCB1 algorithm in \cite{jingxuan2021}.}
    \label{fig:dist_UCB1_vs_previous_alg}
\end{figure}

In all simulations we consider $M=5$ arms with homogeneous rewards drawn from normal distributions (truncated to $[0,1]$) with means $[0.6, 0.5, 0.4, 0.3, 0.2]$. Each simulation is run for $T=10,000$ steps on a newly generated $N=20$ agent undirected and connected (Fig.~\ref{fig:dist_UCB1_vs_single_UCB1} - Fig~\ref{fig:dist_UCB1_vs_previous_alg}) or directed and strongly connected (Fig~\ref{fig:directed_normal}) graphs.  The regret curves are obtained by averaging the largest regret of all agents (the regret of the worst-performing agent) over all 100 simulation runs. The colored region represents the error bar, indicating the range of the regret of the worst-performing agent across all runs. For additional simulations on other reward distributions or agent-wise performance comparisons, see Subsection~\ref{appendix:simulation}.

\begin{figure}[!tb]
    \centering
    \includegraphics[width=0.8\textwidth]{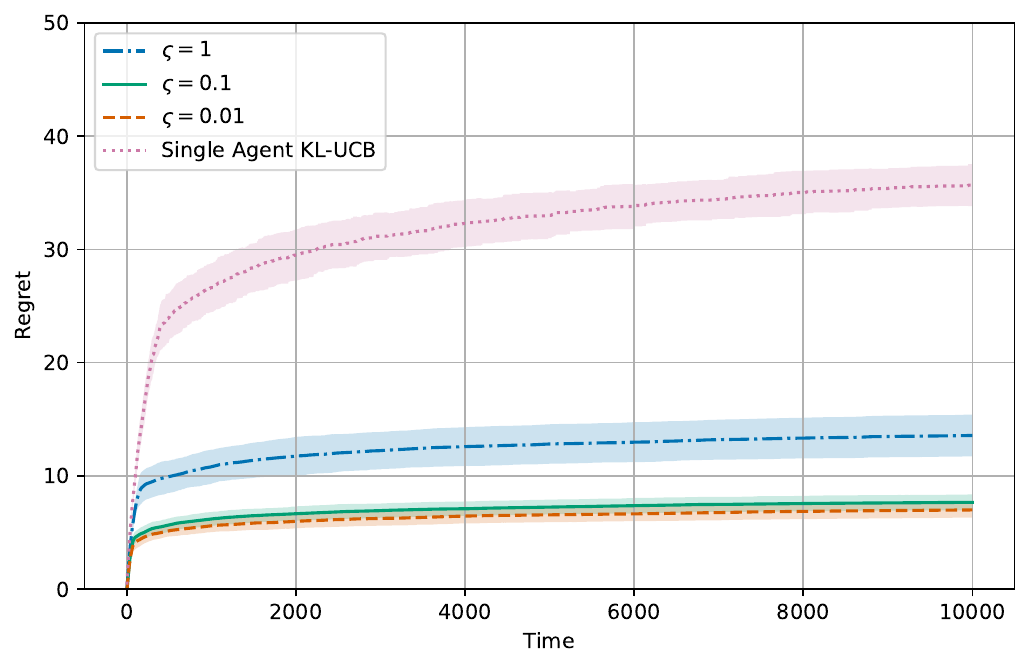}
    \caption{KL-UCB vs. decentralized KL-UCB}
    \label{fig:dist_KL_vs_single_KL}
\end{figure}

   



\begin{figure}[!tb]
    \centering
    \includegraphics[width=0.8\textwidth]{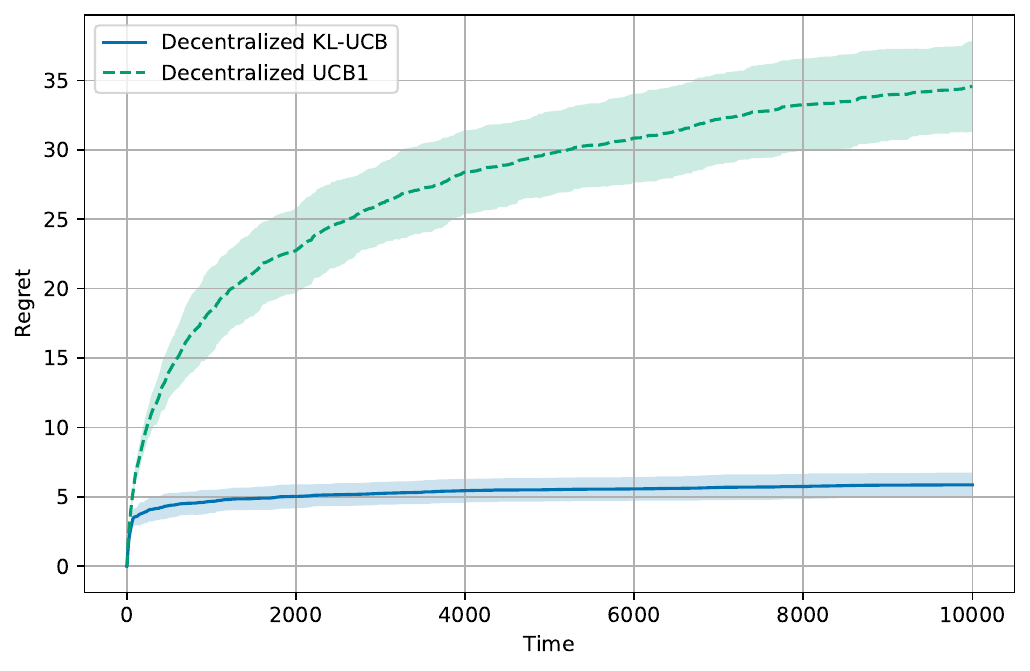}
    \caption{Plots of the mean regrets of decentralized KL-UCB vs. decentralized UCB1 for $\varsigma=\beta=0.01$}
    \label{fig:dist_KL_vs_dist_UCB1}
\end{figure}

\textbf{Observations:}
From Figs.~\ref{fig:directed_normal} and~\ref{fig:dist_UCB1_vs_single_UCB1} we immediately see that the decentralized versions of UCB1 perform better than their single-agent counterparts with suitable choice of parameter $\beta$ over both undirected and connected graphs and directed and strongly connected graphs. Similar performance pattern is also observed in Fig.~\ref{fig:dist_KL_vs_single_KL} for comparison between our decentralized versions of KL-UCB and its single-agent counterpart.
Besides, it appears that as a general pattern, a lower $\varsigma$ (or $\beta$) corresponds to a lower regret, with diminishing returns as the parameter approaches zero. These two observations match the theoretical results in Remark~\ref{re:outperformKL} and Remark~\ref{re:outperformUCB1},~respectively. 

Fig.~\ref{fig:dist_KL_vs_dist_UCB1} highlights a significant outperformance of decentralized KL-UCB over decentralized UCB1 when both parameters are chosen as 0.01. This is consistent with the well-known single-agent case, namely KL-UCB outperforms UCB1 \cite{KLUCB}. Note that from Remark~\ref{re:outperformKL} and Remark~\ref{re:outperformUCB1}, sufficiently small parameter values, like
0.01, lead to a nearly optimal performance for both decentralized algorithms. 
Fig.~\ref{fig:dist_UCB1_vs_previous_alg} shows the outperformance of our decentralized UCB1 algorithm over the decentralized UCB1 algorithm in \cite{jingxuan2021}, which is the only fully decentralized UCB1 algorithm in the existing literature. It can be seen that the outperformance is significant when $\beta$ is chosen as a small value (e.g. as small as $0.1$).

\subsection{Additional Simulations}\label{appendix:simulation}
\begin{figure}[!tb]
    \centering
    \includegraphics[width=0.8\textwidth]{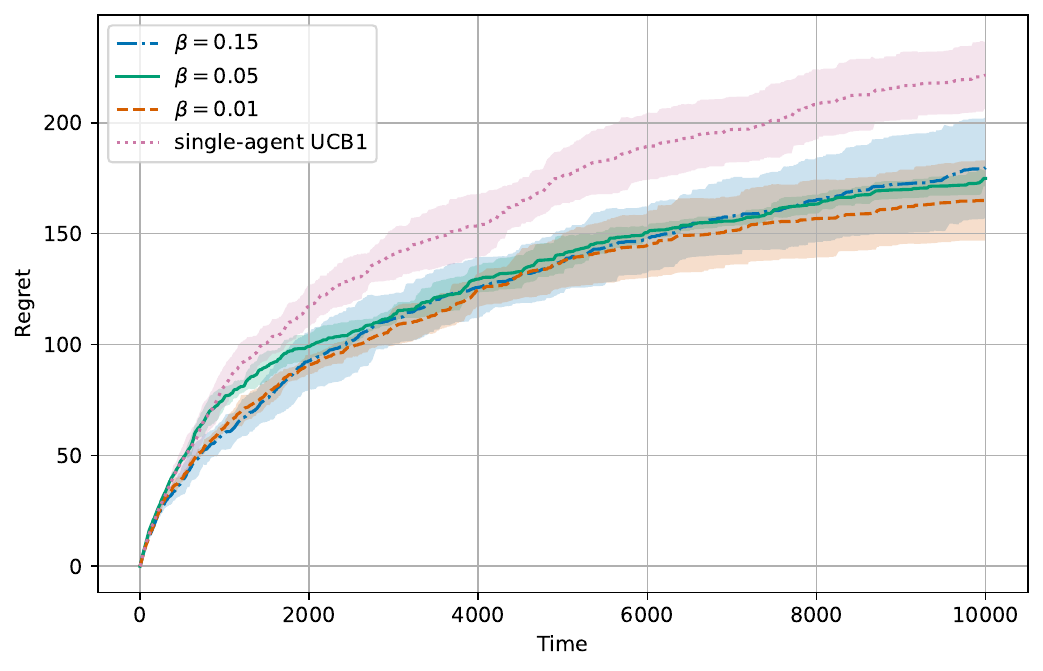}
    \caption{A plot of the mean regrets of UCB1 and decentralized UCB1 over directed graphs for various $\beta$, considering beta distributions for arm rewards.}
    \label{fig:directed_beta}
\end{figure}

\begin{figure}[!tb]
    \centering
    \includegraphics[width=0.8\textwidth]{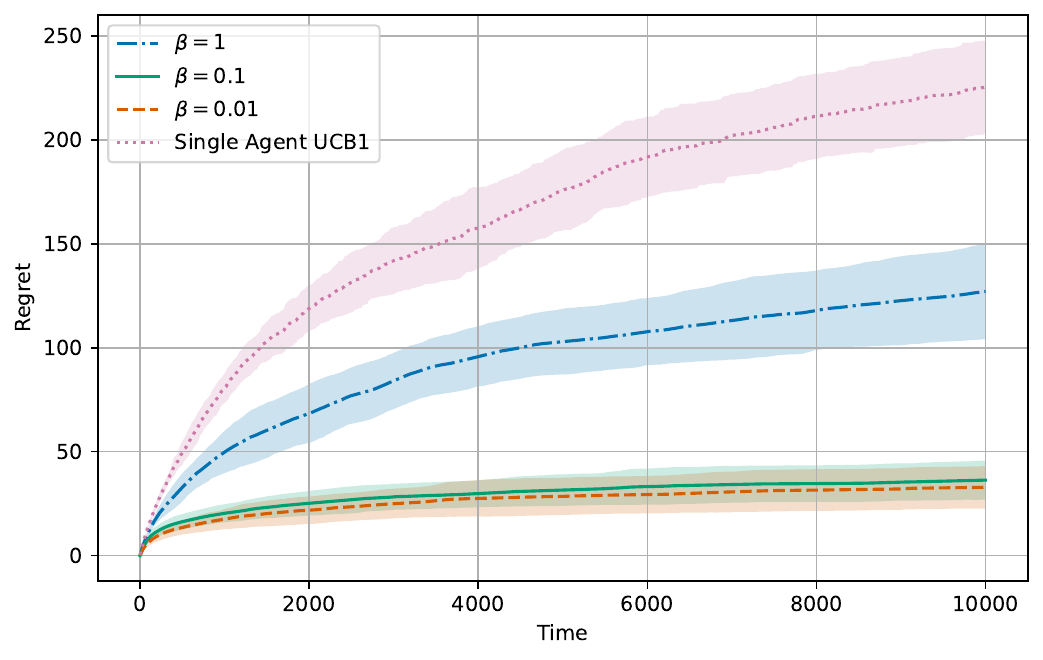}
    \caption{A plot of the mean regrets of UCB1 and decentralized UCB1 for various $\beta$, considering beta distributions for arm rewards.}
    \label{fig:Dist_Single_UCB1_beta}
\end{figure}

\begin{figure}[!tb]
    \centering
    \includegraphics[width=0.8\textwidth]{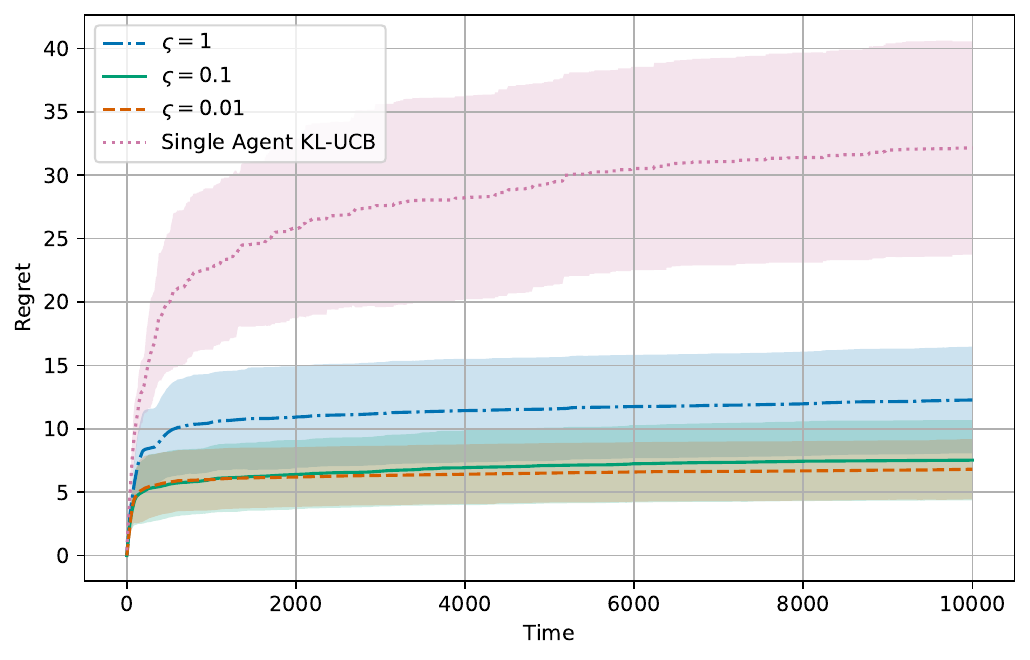}
    \caption{A plot of the mean regrets of KL-UCB and decentralized KL-UCB for various $\varsigma$, considering beta distributions for arm rewards.}
    \label{fig:Dist_Single_KL_UCB_beta}
\end{figure}

We first repeat the simulations from the main paper, except now using beta distributions with means [0.6, 0.5, 0.4, 0.3, 0.2] for arm rewards. Regrets are averaged over a total of 100 runs for $T=10000$ steps each, with a unique $N=20$ agent graph $\bbb G$ generated for each run.
Fig.~\ref{fig:directed_beta} displays the regrets for UCB1 and decentralized UCB1 for various values of $\beta$ over directed and strongly connected graphs. Fig.~\ref{fig:Dist_Single_UCB1_beta} illustrates the regrets for UCB1 and decentralized UCB1 for various values of $\beta$ over undirected and connected graphs.
Fig. \ref{fig:Dist_Single_KL_UCB_beta} presents the regrets 
for KL-UCB and decentralized KL-UCB for various values of $\varsigma$.


\begin{figure}[!tb]
    \centering
    \includegraphics[width=0.8\textwidth]{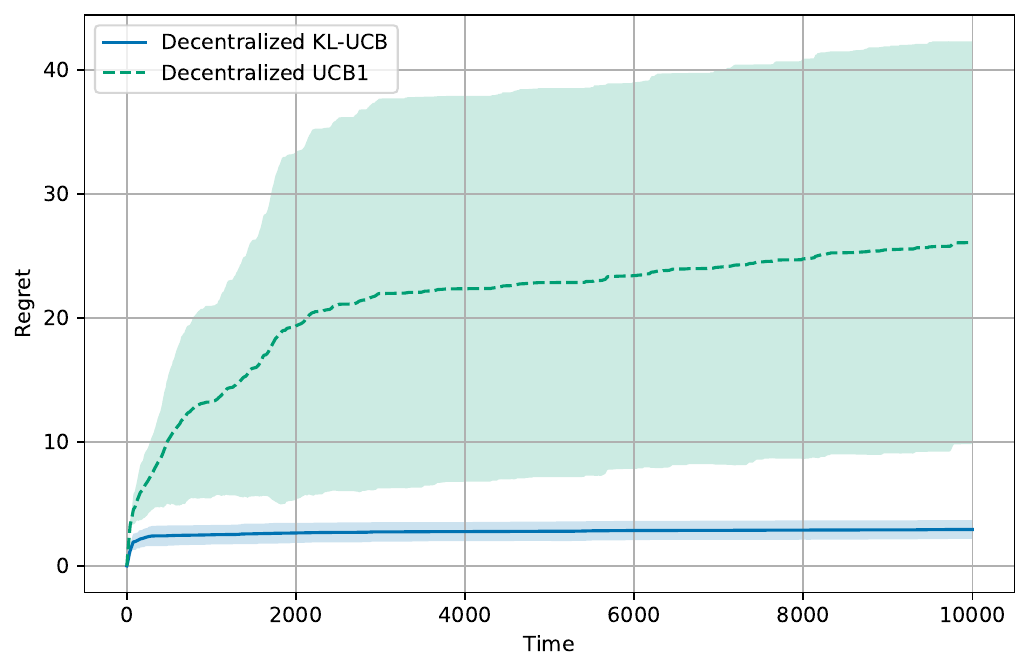}
    \caption{Plots of the mean regrets of decentralized KL-UCB vs. decentralized UCB1 for $\varsigma=\beta=0.01$, considering beta distributions for arm rewards.}
    \label{fig:Dist_UCB1_vs_Dist_KL_UCB_beta}
\end{figure}

\begin{figure}[!tb]
    \centering
    \vspace{-.2in}    
    \includegraphics[width=0.7\textwidth]{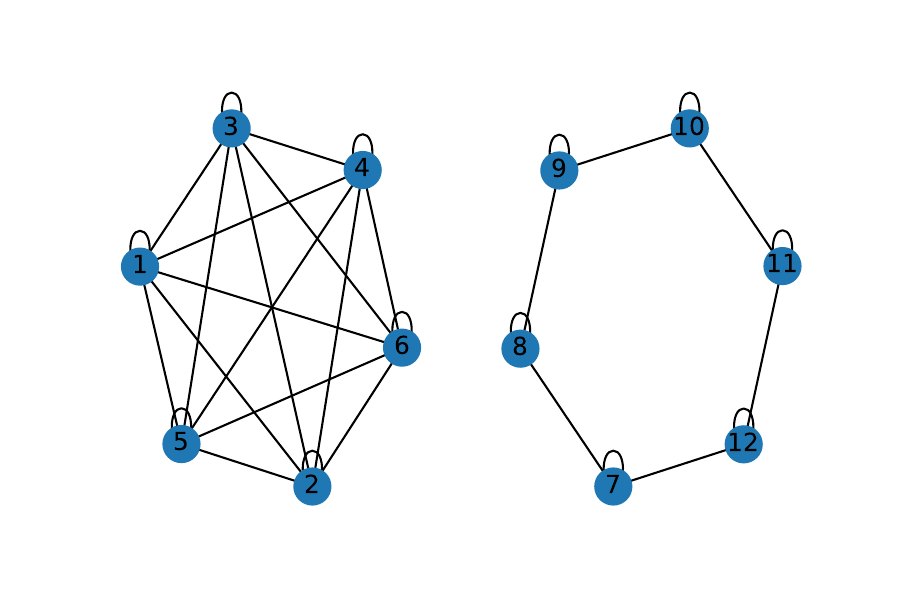}
    \vspace{-.3in}
    \caption{The $N=12$ graph used for the simulations.}
    \label{fig:small_graph}
\end{figure}

\begin{figure}[!tb]
    \centering
    \includegraphics[width=0.95\textwidth]{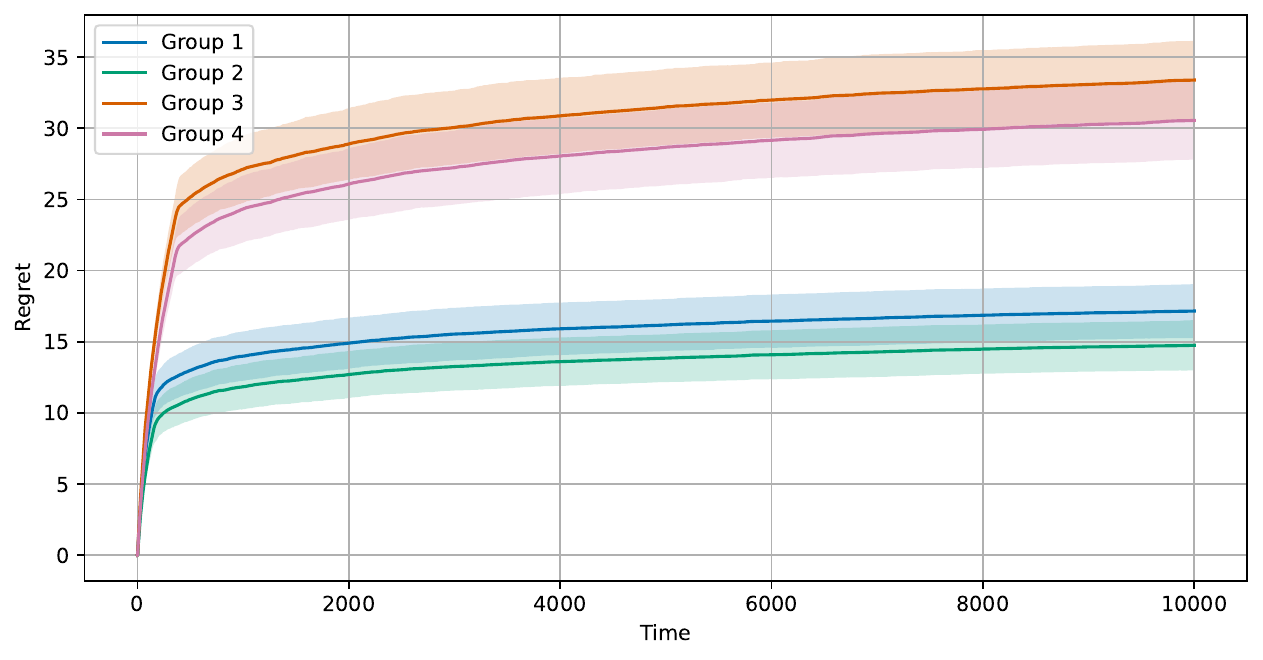}
    \caption{Plots of each group's averaged regret of decentralized KL-UCB over the graph illustrated in Fig.~\ref{fig:small_graph}.}
    \label{fig:KL_UCB_small_graphs_1}
\end{figure}

\begin{figure}[!tb]
    \centering
    \includegraphics[width=0.95\textwidth]{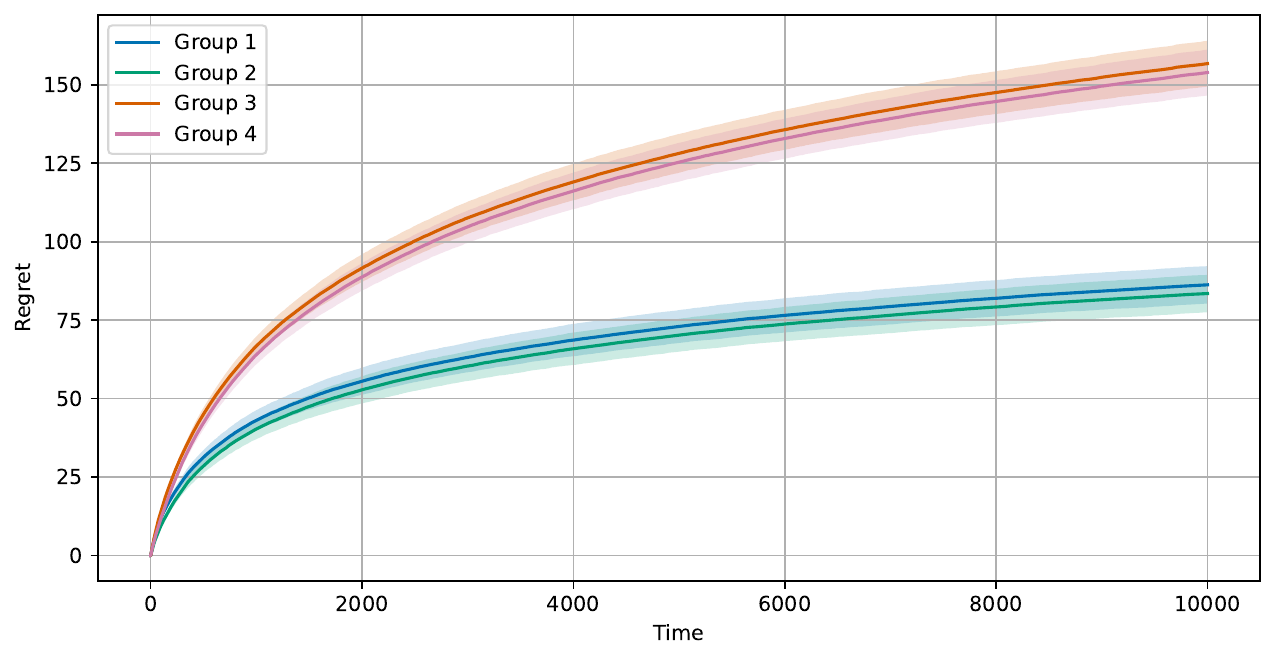}
    \caption{Plots of each group's averaged regret of decentralized UCB1 over the graph illustrated in Fig.~\ref{fig:small_graph}.}
    \label{fig:UCB1_small_graphs_1}
\end{figure}

Notice from Algorithms~\ref{algorithm:KL} and \ref{algorithm:UCB1} that agents can have separate parameter designs and they may have different number of neighbors. To illustrate the effects of $\varsigma_i$ (or $\beta_i$) and $|\scr N_i|$ on each agent's performance for both decentralized algorithms, we consider a twelve-agent undiretced and unconnected graph as shown in Fig. \ref{fig:small_graph}. This graph is composed of two connected components: a six-agent complete graph and a six-agent cycle graph. As described in Section~\ref{sec:problemformulate}, each vertex has a self-loop, indicating that each agent takes itself as a neighbor. 
We divide the agents into four groups, with agents 1, 2, 3 in Group 1, agents 4, 5, 6 in Group 2, agents 7, 8, 9 in Group 3, and agents 10, 11, 12 in Group 4, and set $\varsigma_i$ (or $\beta_i$) at 1 for agents in Groups 1 and 3, and at 0.01 for agents in Groups 2 and 4. Then, the (parameter, number of neighbor) pairs for Group 1-4 are $(1, 6), (0.01, 6), (1, 3), (0.01, 3)$ respectively. We consider $M=5$ arms with rewards drawn from normal distributions truncated to $[0,1]$ with means $[0.6, 0.5, 0.4, 0.3, 0.2]$. We average the regrets for each agent over 100 trials, with each trial ran for $T=10000$ steps. We plot the averaged regret taken across all agents in each group for both Algorithm~\ref{algorithm:KL} and Algorithm~\ref{algorithm:UCB1}, as shown in Fig. \ref{fig:KL_UCB_small_graphs_1} and \ref{fig:UCB1_small_graphs_1}, respectively.

\textbf{Observations:}
Observations consistent with Section~\ref{sec:simulations} can be taken from Fig.~\ref{fig:directed_beta} - \ref{fig:Dist_UCB1_vs_Dist_KL_UCB_beta} for simulations on beta reward distributions, which again validate our theoretical results. As for Fig.~\ref{fig:KL_UCB_small_graphs_1} and \ref{fig:UCB1_small_graphs_1}, it is easy to see that the regret satisfies: Group 3 $>$ Group 4 $>$ Group 1 $>$ Group 2 for both decentralized algorithms, which indicates that for each agent $i,$ a lower $\varsigma_i$ (or $\beta_i$) and a larger size of its neighbor set correspond to a better performance. This observation is in general consistent with the regret bound in both Theorem~\ref{thm:KL} and Theorem~\ref{thm:editedmaintheorem}. Moreover, taking Group 1 and 2 in Fig.~\ref{fig:UCB1_small_graphs_1} as an example, the only difference between the two groups is the agents' choice of parameter, with one being 1 and the other being 0.01. The difference in performance of the two groups is far less than that in Fig.~\ref{fig:dist_UCB1_vs_single_UCB1} or Fig.~\ref{fig:Dist_Single_UCB1_beta}, where we assume an identical parameter for each agent. This indicates that via information fusion, agents influence each other in the exploration process, which may actively prevent the polarization of their performance despite a possible gap in the value of the parameter. Although this conclusion is beyond the scheme of the current paper, we may consider it as a future direction to further improve the regret bound, especially for finite-time analysis. 



\begin{figure}[!tbh]
    \centering
    \includegraphics[width=0.9\textwidth]{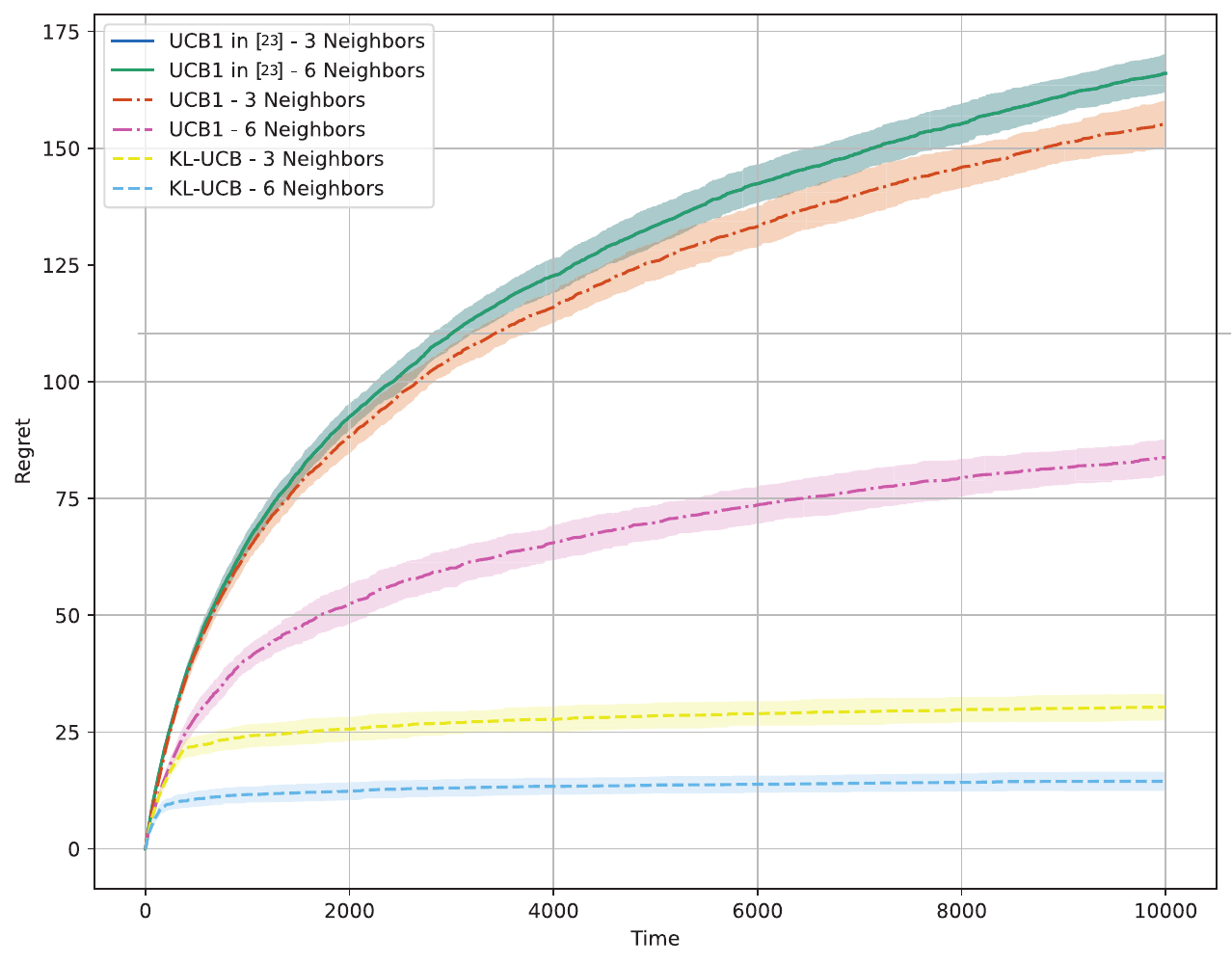}
    \caption{Plots of the averaged regret of decentralized UCB, UCB1, and KL-UCB over the graph illustrated in Fig.~\ref{fig:small_graph}. $\beta$ and $\varsigma$ are fixed to 0.01 for each agent in decentralized UCB1 and decentralized KL-UCB, respectively. (UCB1 in \cite{jingxuan2021} - 3 Neighbors and UCB1 in \cite{jingxuan2021} - 6 Neighbors overlap.)}
    \label{fig:algorithm_comparison}
\end{figure}

\textbf{Performance Comparison:}
To experimentally compare the performance of the decentralized UCB1 algorithm presented in \cite{jingxuan2021} 
and the decentralized UCB1 and KL-UCB algorithms proposed in this paper, we again use the neighbor graph given in Fig.~\ref{fig:small_graph}. We consider $M=5$ arms with rewards drawn from normal distributions truncated to $[0,1]$ with means $[0.6, 0.5, 0.4, 0.3, 0.2]$. Both $\beta$ and $\varsigma$ are set as 0.01 for each agent. 
For each algorithm, we average the regrets for each agent over 10 trials, with each trial being run for $T=1000$ steps, and then average those regrets over all those agents which have the same number of neighbors. These regrets are plotted in Fig.~\ref{fig:algorithm_comparison}.

There are two key factors to take note of from this comparison. First, the decentralized UCB1 algorithm proposed in this paper outperforms the algorithm presented in \cite{jingxuan2021}. This is particularly noticeable as the number of neighbors increases, illustrating how the regret of our decentralized UCB1 algorithm is influenced by an agent's number of neighbors, whereas it is not the case for the algorithm in \cite{jingxuan2021} (so the corresponding two curves appear to overlap in the figure). Second, more importantly, the regret of our decentralized KL-UCB outperforms both decentralized UCB1 algorithms, and also improves as the number of an agent's neighbors increases. This outperformance is expected as KL-UCB is a state-of-the-art bandit algorithm. The simulation thus validates that our decentralized KL-UCB algorithm has the best performance so far in terms of regret bounds.

\section{Conclusion}\label{sec:conclusion}

In this paper, we have studied a decentralized multi-armed bandit problem over undirected graphs and proposed fully decentralized version of KL-UCB and UCB1, which provably achieve a better logarithmic asymptotic regret than their single-agent counterparts. 
The proposed decentralized bandit design framework is expected to be applicable to other UCB type algorithms. 
While theoretically we only focus on the asymptotic regret, finite-time simulations also show the same pattern. 
Future directions are to study the limitations of the paper, including better theoretical result for finite-time analysis, and analysis on directed graphs.

\bibliographystyle{unsrt}
\bibliography{1,2,3,distributedMAB,pylibs}



\newpage

\section*{Appendix: Pseudocode}\label{sec:pseudo}

\vspace{.2in}

\begin{algorithm2e}[!th]
\LinesNumbered
\DontPrintSemicolon   
\caption{Decentralized KL-UCB}
\label{algorithm:KL}
\SetKw{KwRet}{Return}
\KwIn{$\bbb G, T, Q_i(t)$}
\textbf{Initialization} Each agent samples each arm exactly once. Initialize $z_{i,k}(0)=\bar{x}_{i,k}(0)=X_{i,k}(0)$, $m_{i,k}(0)=n_{i,k}(0)=1$
\BlankLine
\For{$t = 0,\ldots,T$}{
$\scr A_{i}=\emptyset$ \;
\For{$k=1,\ldots,M$}{
\If{$n_{i,k}(t)\leq m_{i,k}(t)-M$} {Agent $i$ puts $k$ into a set $\scr A_{i}$}}
\eIf{$\scr A_{i}= \emptyset$}{
$a_i(t+1)=\argmax_k\max\{q:n_{i,k}(t)d(z_{i,k}(t);q)\le Q_i(t)\}$ \hfill   \tcp*{optimal arm in belief}}{
$a_{i}(t+1)$ is randomly chosen from $\scr A_{i}$ \hfill  \tcp*{ for exploration consistency}}
Agent $i$ sends $m_{i,k}(t)$ and $z_{i,k}(t), \forall k \in [M],$ to $j\in \scr{N}_i$ \hfill  \tcp*{ information propagation}
Agent $i$ receives $m_{j,k}(t), z_{j,k}(t), \forall k \in [M],$ from $j\in \scr{N}_i$ \;
$n_{i,k}(t+1)=n_{i,k}(t),\;\forall k\in [M]$ \tcp*{information updating}
$n_{i,a_{i}(t+1)}(t+1)=n_{i,a_{i}(t+1)}(t)+1$ \;
$z_{i,k}(t+1)=\sum_{j\in\scr{N}_i}w_{ij}(t)z_{j,k}(t)+\bar{x}_{i,k}(t+1)-\bar{x}_{i,k}(t)$\;
$m_{i,k}(t+1)= \;\max \{n_{i,k}(t+1),\;\max_{j\in\scr N_i}m_{j,k}(t)\}$   \;
}
\BlankLine
\end{algorithm2e}

\begin{algorithm2e}[!th]
\LinesNumbered
\setcounter{AlgoLine}{0}
\DontPrintSemicolon   
\caption{Decentralized UCB1}
\label{algorithm:UCB1}
\SetKw{KwRet}{Return}
\KwIn{$\bbb G, T, C(t,n_{i,k}(t))$}
\textbf{Initialization} Each agent samples each arm exactly once. Initialize $z_{i,k}(0)=\bar{x}_{i,k}(0)=X_{i,k}(0)$, $m_{i,k}(0)=n_{i,k}(0)=1$
\BlankLine
\For{$t = 0,\ldots,T$}{
$\scr A_{i}=\emptyset$ \;
\For{$k=1,\ldots,M$}{
\If{$n_{i,k}(t)\leq m_{i,k}(t)-M$} {Agent $i$ puts $k$ into a set $\scr A_{i}$}}
\eIf{$\scr A_{i}= \emptyset$}{
$a_i(t+1)=\argmax_k\{z_{i,k}(t)+ C(t,n_{i,k}(t))\}$\hfill  \tcp*{ optimal arm in belief}}{
$a_{i}(t+1)$ is randomly chosen from $\scr A_{i}$\hfill  \tcp*{ for exploration consistency}}
Agent $i$ sends $m_{i,k}(t)$ and $z_{i,k}(t), \forall k \in [M],$ to $j\in \scr{N}_i$ \hfill  \tcp*{ information propagation}
Agent $i$ receives $m_{j,k}(t), z_{j,k}(t), \forall k \in [M],$ from $j\in \scr{N}_i$ \;
$n_{i,k}(t+1)=n_{i,k}(t),\;\forall k\in [M]$ \hfill\hfill  \tcp*{ information updating}
$n_{i,a_{i}(t+1)}(t+1)=n_{i,a_{i}(t+1)}(t)+1$\;
$z_{i,k}(t+1)=\sum_{j\in\scr{N}_i}w_{ij}(t)z_{j,k}(t)+\bar{x}_{i,k}(t+1)-\bar{x}_{i,k}(t)$\;
$m_{i,k}(t+1)= \;\max \{n_{i,k}(t+1),\;\max_{j\in\scr N_i}m_{j,k}(t)\}$\;
}
\BlankLine
\end{algorithm2e}

\end{document}